\documentclass[final,5p,times,twocolumn]{elsarticle}

\usepackage{graphicx}
\graphicspath{{./}}
\DeclareGraphicsExtensions{.pdf,.png,.jpg,.jpeg}
\usepackage{booktabs}
\usepackage{array}
\usepackage{multirow}
\usepackage{amsmath,amssymb}
\usepackage{xcolor}
\usepackage{float}
\usepackage{subcaption}  
\usepackage[hidelinks]{hyperref}
\usepackage{pgfplots}
\usepackage{ragged2e}

\makeatother
\pgfplotsset{compat=1.18}

\usepackage[font=normalsize]{caption}

\begin{document} 

\begin{frontmatter}

\title{DefoliationCAM: An Edge-AI System for Real-Time Insect Defoliation
Severity Assessment in Grapevines}

\author[a,b]{Fuchen Li}
\author[a,b]{Wei-Zhen Liang\corref{cor1}}
\ead{wei-zhen.liang@unl.edu}
\cortext[cor1]{Corresponding author}
\author[b,c]{Pin-Chu Lai}
\author[a,b]{Xin Qiao}
\author[d]{Jingqiu Chen}
\author[e]{Haw Yen}
\author[a,b]{Zonglin Yang}
\author[a,b]{Chi-En Chiang}
\author[b]{Rudolph Avalos}
\author[b]{Abbigail L. Rath}

\affiliation[a]{organization={Department of Biological Systems Engineering,
University of Nebraska--Lincoln}, city={Lincoln}, state={NE}, country={USA}}
\affiliation[b]{organization={Panhandle Research and Extension Center,
University of Nebraska--Lincoln}, city={Scottsbluff}, state={NE}, country={USA}}
\affiliation[c]{organization={Department of Entomology, University of
Nebraska--Lincoln}, city={Lincoln}, state={NE}, country={USA}}
\affiliation[d]{organization={Department of Biological Systems Engineering, Florida A\&M University}, city={Tallahassee}, state={FL}, country={USA}}
\affiliation[e]{organization={USDA-ARS, National Soil Erosion Research laboratory}, city={West Lafayette}, state={IN}, country={USA}}

\begin{abstract}
Insect-induced leaf defoliation threatens grapevine health and yield, and its
severity is still heavily rely on subjective manual scouting. This study developed a solar-powered edge AI system DefoliationCAM (D-CAM) which estimates the degree of defoliation in grape canopy in vineyard and transmits results through LoRa to a remote web dashboard. The two-stage pipeline first uses YOLOX-Nano to localize the grapevine cordon in a 4K image, and then uses a MobileNet-based classifier enhanced with lightweight channel attention to classify $320\times320$ patches. The 29{,}674 field-collected patches were originally annotated as background or one of four damage levels: no to negligible, mild, moderate, and severe. Merging the original moderate and severe classes reduced the class imbalance ratio from approximately 40:1 to 8.8:1, and synthetic augmentation further increased macro-F1 from 0.768 to 0.838. On the same test set, the selected classifier achieved a higher macro-F1 than ResNet-50 despite having substantially fewer parameters (1.08 M versus 23.52 M). Patch predictions were aggregated into a canopy-level Non-linear Global Defoliation Severity (NGDS) score, validated by an entomologist evaluating 86 held-out canopy images and assigning them to three management categories: (1) Negligible damage, (2) Mild damage warranting continued monitoring, and (3) Action-level damage warranting on-site scouting and insecticide treatment consideration. Quadratic discriminant analysis identified NGDS boundaries of 9\% and 29\% corresponding to category (2) and (3), respectively. The 29\% boundary was consistent with the action threshold range of  25\%--30\% for Japanese beetle management in grapevine. On an NVIDIA Jetson Orin Nano, the deployed pipeline processed each 4K image in approximately 1.5~s using TensorRT FP16, while the complete operating cycle required approximately 57--60~s. A five-week deployment of two field units demonstrated the feasibility of autonomous solar-powered operation, LoRa communication, and remote vineyard monitoring for defoliation. The D-CAM provides a practical platform for frequent, objective, and remote assessment of canopy damage, supporting more targeted vineyard scouting and timely pest-management decisions.
\end{abstract}


\end{frontmatter}

\section{Introduction}

Grapevines (\textit{Vitis} spp.), including wine and table grape cultivars, represent an economically important specialty crop in the United States. In 2025, U.S. vineyards comprised approximately 883,000 bearing acres and produced 5.23 million tons of utilized grapes valued at \$5.83 billion; 882,100 tons were marketed as table grapes, while 4.35 million tons were processed for wine, juice, raisins, and other products~\cite{usda2026}. Maintaining a healthy canopy is essential because leaves supply the photosynthates required for vine growth, berry development, and fruit ripening. Excessive leaf loss can slow sugar accumulation, affect fruit composition and vine growth, and cause multi-season yield losses. 

Japanese beetle (\textit{Popillia japonica} Newman) is an important defoliating pest of grapevines in the U.S. Adult beetles feed on tissue between the major leaf veins causing skeletonization and severe defoliation. Although no single defoliation threshold applies to all grape cultivars, mature bunch grapevines can generally tolerate less than 20\% leaf-area loss, with tolerance varying by cultivar and vine condition~\cite{hammons2010}. In ‘Frontenac’, a cold-hardy wine grape cultivar, fruit quality began to decline when leaf-area loss exceeded approximately 30–35\%, leading to a proposed action threshold of 25–30\% defoliation for Japanese beetle management~\cite{ebbenga2022}. Consistent with these findings, the University of Wisconsin–Madison recommends approximately 30\% defoliation as a practical threshold for considering insecticide treatment, while vines with less damage should continue to be monitored and scouted~\cite{guedot2019}. Continuous monitoring of canopy defoliation is therefore essential for precision viticulture
and early pest intervention. In current practice, this monitoring is performed by manual visual scouting, which is labor-intensive, subjective, and difficult to repeat constantly across large or remote vineyards.

Computer vision, particularly convolutional neural networks, is widely used to identify plant diseases and stress from images, and its development has been driven largely by public datasets~\cite{ferentinos2018,mohanty2016,too2019}. PlantVillage provides
54{,}306 leaf images captured under controlled background and lighting
conditions~\cite{hughes2015plantvillage}. Models trained and evaluated on these relatively uniform images achieved high classification accuracy, but their performance declined when tested on images captured under field conditions. PlantDoc addressed this limitation by providing 2,598 non-laboratory images covering 13 plant species and 17 diseases~\cite{singh2020plantdoc}, while the Plant Pathology 2020 dataset included 3,651 images of apple leaves covering three diseases collected under varying illumination, viewing angles, and surfaces and image noise~\cite{thapa2020}. PlantDoc, however, is a small, web-scraped collection of mixed image quality, whereas Plant Pathology 2020 is of higher quality and field-collected but limited to apple foliar diseases. Although these datasets better represent field conditions, they are considerably smaller than PlantVillage and were developed primarily for disease recognition from individually collected images rather than long-term, continuous monitoring of crops in the field. Kamilaris and Prenafeta-Bold\'u~\cite{kamilaris2018} similarly noted that many agricultural deep-learning studies relied on relatively small datasets and evaluated models using images from the same dataset, leaving their performance under different field conditions uncertain. Although computer vision has been used in disease recognition, long-term field image datasets that captured change in illumination, canopy, soil backgrounds, viewing angles, and leaf occlusion remain limited. This gap is particularly evident for insect-caused grapevine defoliation, for which few studies~\cite{hernandez2022assessment,hirigoyen2025assessing} have examined automated, canopy-level severity assessment using images collected at regular intervals over field deployments.

Class imbalance is another concern when models are developed from field images, because severe damage may be represented by far fewer samples than healthy or mildly affected conditions. Models trained on imbalanced datasets can favor the larger classes and perform poorly on rare but important cases~\cite{johnson2019survey,krawczyk2016}. Common approaches include minority-class resampling, such as SMOTE~\cite{chawla2002smote}; loss functions designed to emphasize rare or difficult samples, such as focal loss~\cite{lin2017focal} and class-balanced loss~\cite{cui2019class}; and
decoupling representation learning from classifier learning for long-tailed
recognition~\cite{kang2020decoupled}. However, their suitability for grapevine defoliation classification requires further evaluation because images of moderate and severe damage are less common than those showing no to negligible or mild damage, and adjacent severity levels can be visually similar.

Processing images at the edge reduces the need to transmit large, high-resolution files from the field. Raspberry Pi microcomputers have seen widespread adoption in agricultural monitoring due to their compact form factor, low cost, and straightforward integration with cameras and sensors~\cite{talaat2026,chiang2025,yang2026}. 
However, without an external AI accelerator, deep-learning inference relies primarily on the CPU, which limits model complexity and processing speed. Raspberry Pi–based systems have therefore used hand-crafted features and conventional classifiers~\cite{kamath2019}, or lightweight neural
networks~\cite{gonzalez2021}, often at the cost of longer processing times~\cite{yang2026}, or
single-digit frame rates~\cite{zarboubi2024}. In comparison, NVIDIA Jetson platforms provide GPU acceleration for on-device inference and have been used for grape leaf disease classification~\cite{karim2024}, real-time weed detection and target spraying~\cite{assuncao2022}, robotic vineyard pruning~\cite{pacioni2025}, and leaf-disease detection on Ginkgo~\cite{wang2025plfynet} and
peanut~\cite{sun2025}. The NVIDIA Jetson Orin Nano therefore provides a more suitable platform for running computer-vision models on high-resolution vineyard images. To maximize inference efficiency on such edge platforms, lightweight backbones such as MobileNet~\cite{howard2019}, together with channel-attention mechanisms~\cite{hu2018}, offer efficient building blocks for reducing computation while preserving accuracy under such edge constraints.

Moving inference to the edge does not remove the practical constraints of field deployment. Many vineyards lack grid power and reliable cellular service, requiring monitoring devices to operate on solar power and use low-bandwidth communication. Classical agricultural wireless sensor networks that rely on non-vision sensors such as soil moisture, temperature, and humidity have already struggled with limited battery life, unreliable radio links, environmental exposure, and hardware failure over a single growing season~\cite{ojha2015, ruiz2009, langendoen2006}. Image acquisition and deep-learning inference demand orders of magnitude more compute per cycle than these conventional sensors, making power management a critical design consideration for solar-powered vision systems~\cite{tolle2005, jawad2017}. Albanese et al.~\cite{albanese2021pest} demonstrated solar-powered edge inference for insect counting in pheromone traps, but that system operated intermittently within a bounded imaging enclosure containing a handful of target objects per capture. Existing Jetson-based systems have primarily been integrated into drones or robotic platforms powered by rechargeable batteries for short, supervised field operations lasting hours to days~\cite{menshchikov2021,mwitta2024}. Continuous unattended monitoring, however, requires the coordinated design of solar power management, computationally efficient inference, and low-bandwidth long-range communication such as LoRa.

\begin{figure*}[!t]
\centering
\includegraphics[width=\linewidth,trim=0 80 0 0,clip]{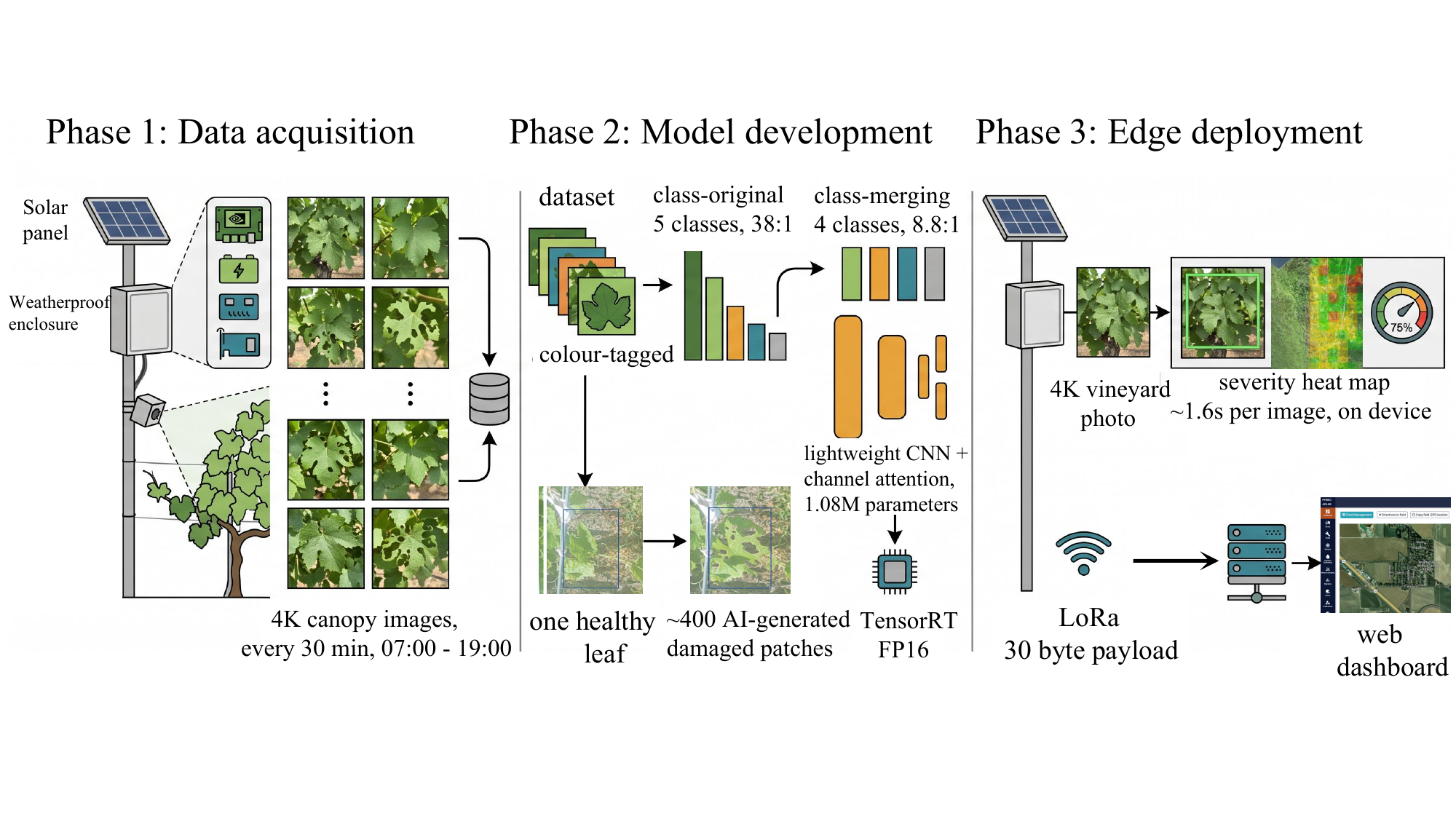}
\caption{Three-phase framework for the development and deployment of D-CAM.
(1) scheduled acquisition of 4K grape-canopy images using a solar-powered field unit; (2) dataset preparation, class consolidation, minority-class augmentation, model training, and TensorRT FP16 conversion; and (3) on-device canopy localization, patch-level severity assessment, and LoRa transmission of results to a web dashboard.}
\label{fig:framework}
\end{figure*}

To address these requirements, this study developed and evaluated DefoliationCAM (D-CAM), a solar-powered edge-AI system for repeated, canopy-level assessment of insect-induced defoliation in grapevines. The main contributions of this study are threefold: (1) D-CAM was developed as a solar-powered edge-AI system that acquires and processes 4K grape-canopy images on an NVIDIA Jetson Orin Nano and transmits severity results through LoRa. The system includes a custom battery-monitoring circuit and software-based time persistence for unattended operation without a battery-backed real-time clock or network time synchronization. (2) A lightweight two-stage pipeline was developed by combining YOLOX-Nano with a 1.08 M-parameter classifier incorporating lightweight channel attention (LCA) and the Non-linear Global Defoliation Severity (NGDS) score. The complete pipeline contains fewer than 2 M parameters, and the final four-class model with synthetic minority-class augmentation achieved 83.6\% accuracy and a macro F1 score of 0.838. (3) The NGDS score and the complete monitoring system were evaluated through an independent assessment by an entomologist and a five-week field deployment. Quadratic discriminant analysis identified three management-oriented categories: no to negligible damage (NGDS< 9\%), mild damage warranting continued monitoring (NGDS= 9–29\%), and action-level damage requiring prompt scouting and consideration of insecticide treatment (NGDS> 29\%). The 29\% NGDS boundary was numerically consistent with the range of 25–30\% and the practical threshold of approximately 30\% for Japanese beetle management in bunch grapes, supporting the agronomic relevance of the NGDS score~\cite{ebbenga2022,guedot2019}.
\section{Materials and Methods}

\subsection{System Overview}

\begin{figure*}[!t]
\centering

\includegraphics[width=0.9\linewidth,trim=0 50 0 0,clip]{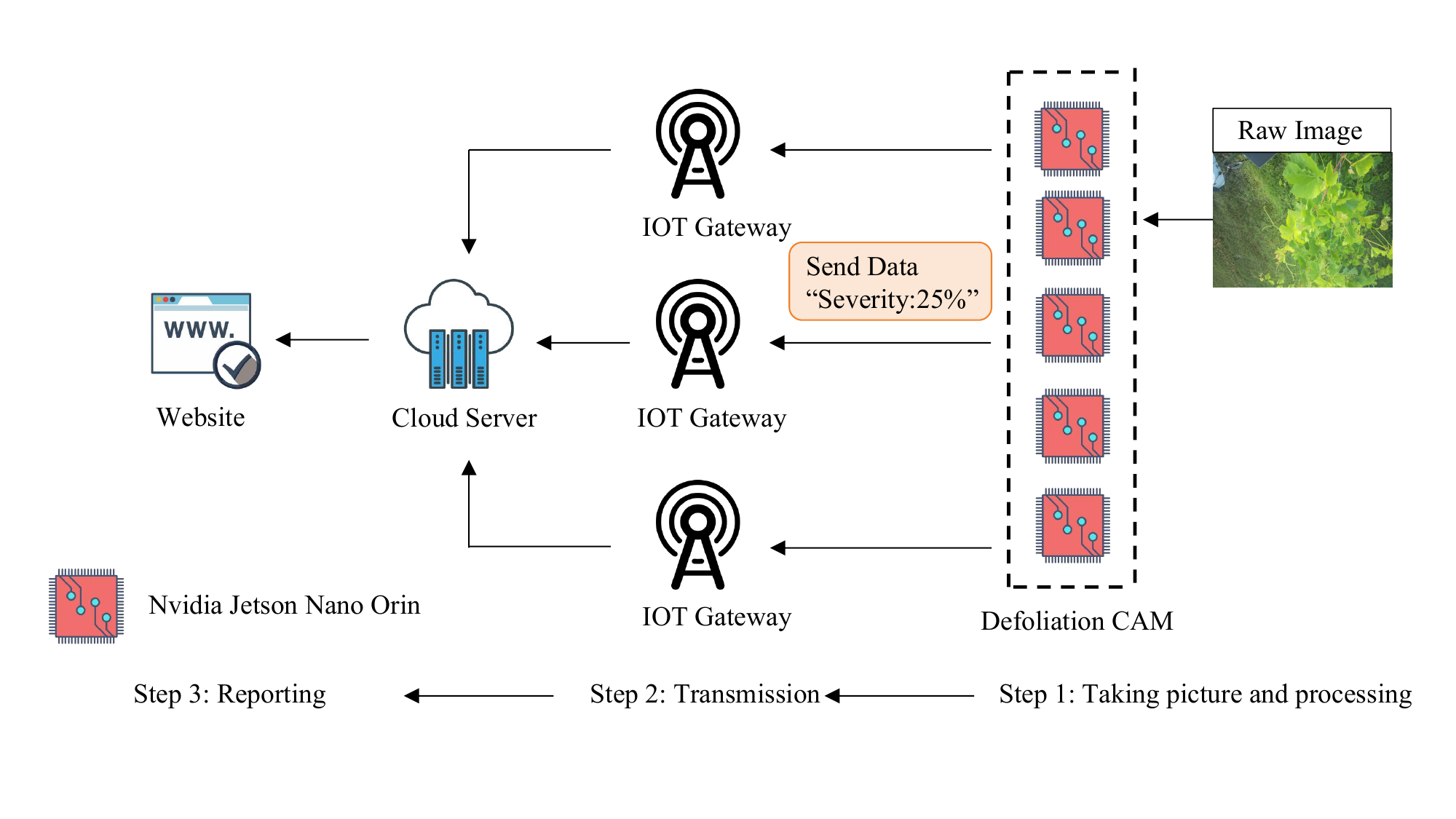}
\caption{System architecture and data flow of D-CAM.}
\label{fig:architecture}

\vspace{1.5em}

\includegraphics[width=0.9\linewidth,trim=70 0 180 0,clip]{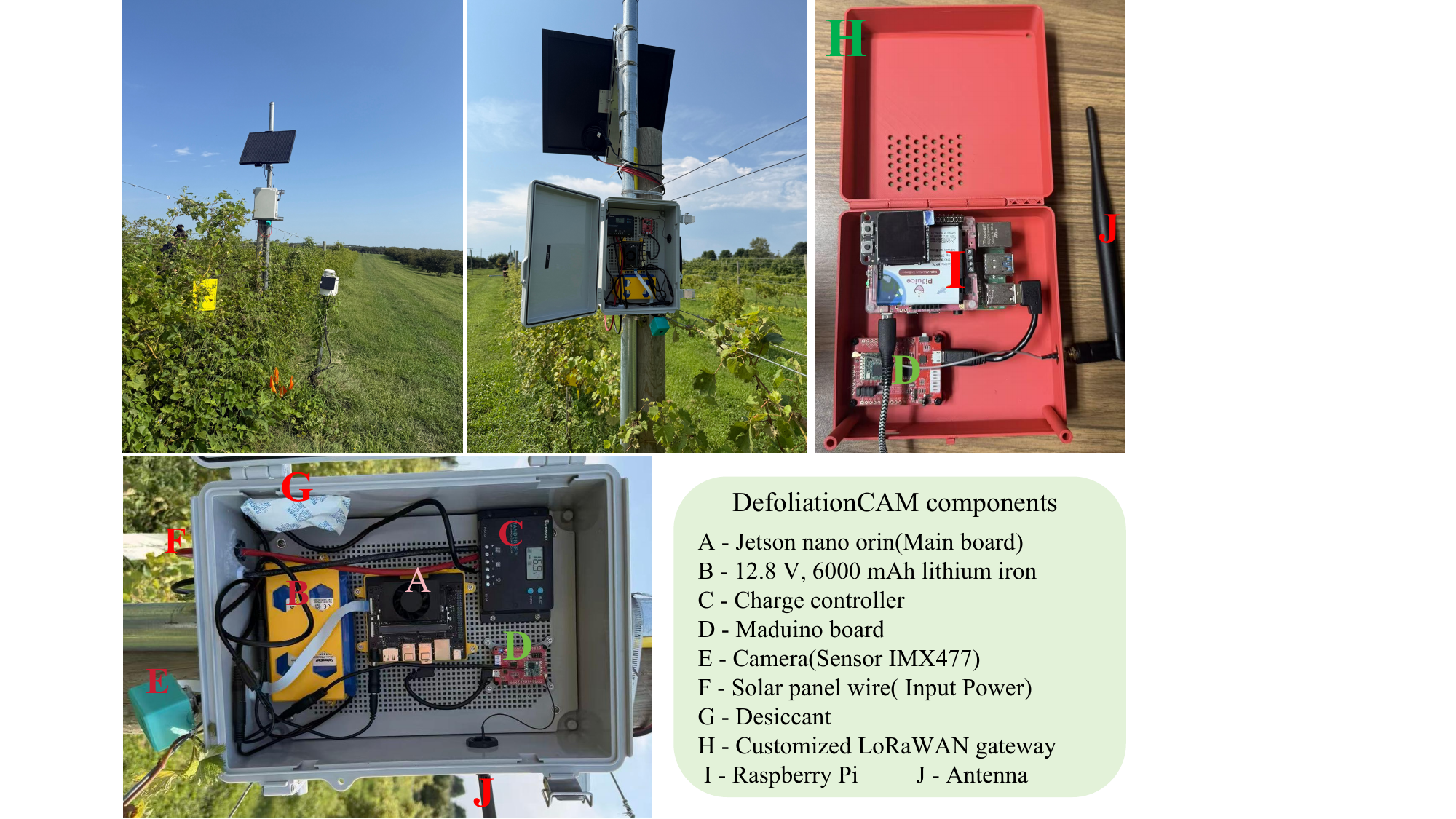}
\caption{Field installation and hardware configuration of D-CAM.}
\label{fig:fieldunit}

\end{figure*}

DefoliationCAM (D-CAM) was developed using a three-phase framework (Fig.~\ref{fig:framework}).
In \emph{Phase~1, data acquisition}, a solar-powered field unit mounted on a vineyard trellis post (Fig.~\ref{fig:fieldunit}) captured 4K grape-canopy images at 30-min sampling intervals under naturally varying illumination, occlusion, weather, and background conditions. In \emph{Phase~2, model development}, canopy regions were manually annotated and tiled into $320\times320$ patches. The two most under-represented levels in the original five-class severity scheme were consolidated into a single \emph{damaged} class, yielding a four-class dataset with a substantially reduced imbalance ratio. Synthetic patches generated with a generative model were then added to further augment the \emph{damaged} class, and the balanced dataset was used to train a lightweight classification network. The trained model was then converted to a TensorRT FP16 engine for on-device inference. In \emph{Phase~3, edge deployment}, the trained model was deployed on the field unit to process newly captured images and generate canopy-level severity scores transmitted over LoRa through a web-based dashboard.

The deployed D-CAM system comprises multiple field units, a custom LoRa gateway, and a cloud server (Fig.~\ref{fig:architecture}). Each field unit operated a scheduled duty cycle. Between 07:00 and 19:00, the unit woke every 30 min to capture an image, read the battery status,  perform inference, and transmit the results. The Jetson was configured in the 15 W power mode during active processing and returned to a low-power sleep state (Linux suspend-to-RAM) after completing each cycle. Following the final cycle at 19:00, the unit remained in this sleep state, with the cooling fan disabled, until automatically resuming at 07:00 the following morning.

At each scheduled cycle, each field unit captures a 4K canopy image and executes the complete two-stage inference pipeline on an NVIDIA Jetson Orin Nano. The device ID,  Unix timestamp, NGDS score, battery state of charge, and operating status are encoded in a 30-byte message and transmitted at 915 MHz using a Maduino LoRa board. The custom gateway, comprising a Raspberry Pi connected to a second Maduino LoRa board and an antenna, receives the message and forwards it over Wi-Fi to the cloud server. The server stores the records and displays the timestamp, NGDS score, and battery state of charge on the web dashboard at \url{https://phrec-irrigation.com}. Because the 4K images are processed on-device and only compact summary data are transmitted, Internet connectivity is required only at the gateway, allowing the field units to operate in vineyard locations without cellular coverage.

\begin{figure}[!t]
\centering
\includegraphics[width=\linewidth]{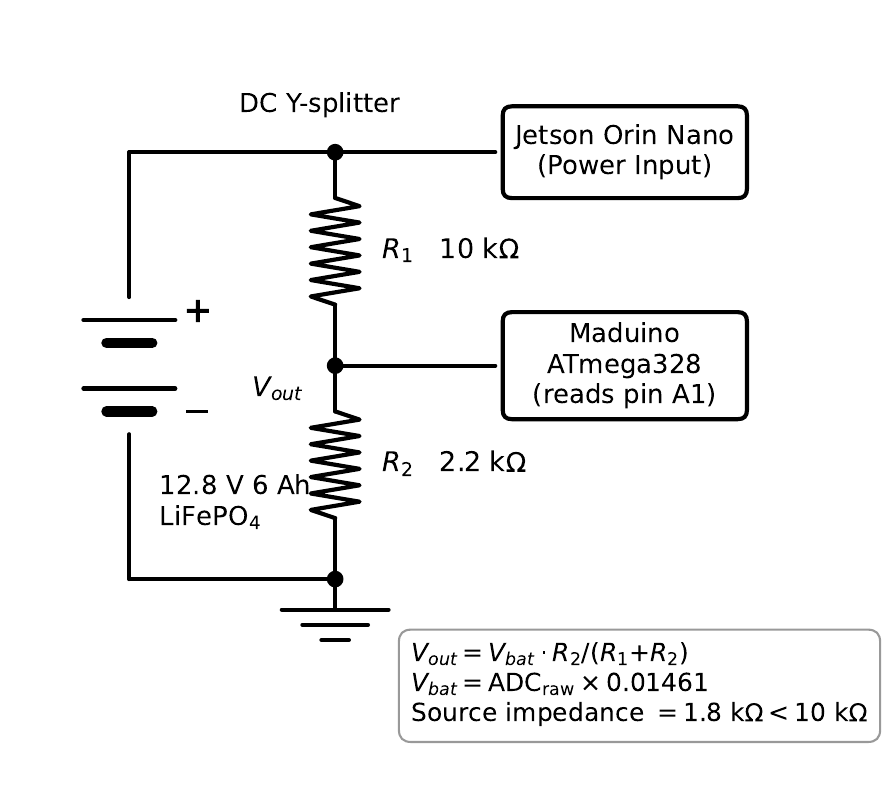}
\caption{Custom analog battery voltage-sensing circuit. A DC Y-splitter taps the TalentCell 12.8\,V battery: one branch powers the Jetson Orin Nano, while the other feeds a resistive divider ($R_1=10$ k$\Omega$, $R_2=2.2$ k$\Omega$) whose midpoint ($V_{out}$) is read on the Maduino (ATmega328) analog pin A1. The divider keeps the source impedance ($\sim$1.8 k$\Omega$) below the 10 k$\Omega$ ATmega328 ADC limit, and the raw reading is converted to battery voltage via $V_{bat} = \mathrm{ADC_{raw}} \times 0.01461$.}
\label{fig:3}
\end{figure}

\begin{figure}[b]
\centering
\includegraphics[width=\linewidth]{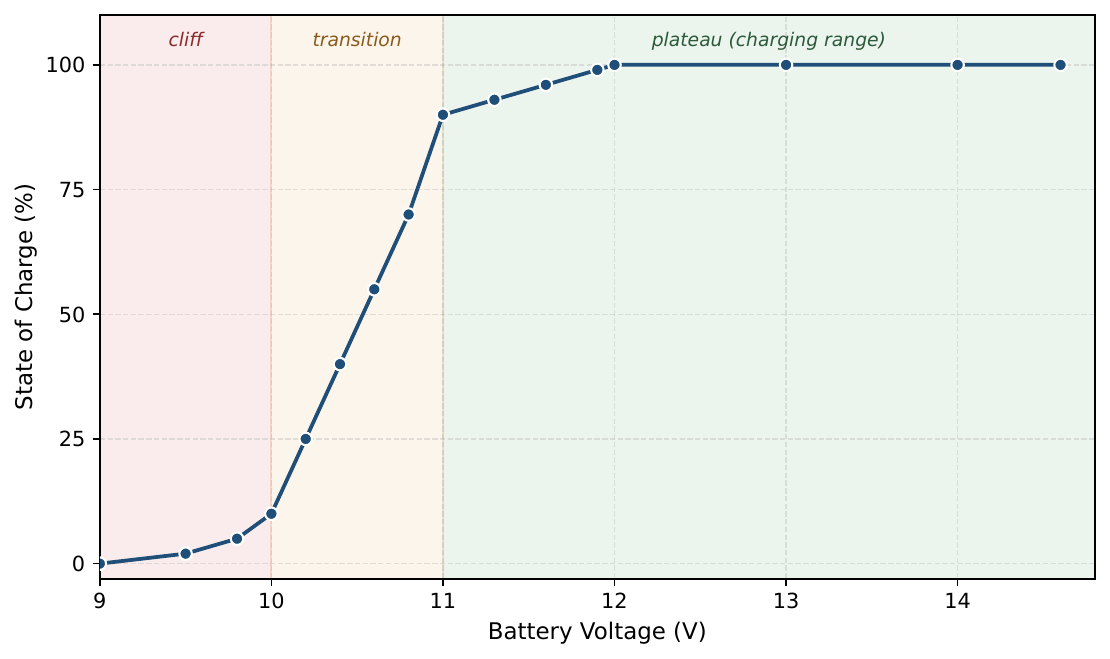}
\caption{ Battery voltage-to-state-of-charge (SOC) calibration curve for the 12.8 V, 6000 mAh LiFePO$_4$ battery. SOC was estimated by linear interpolation between the calibration points, with voltage readings of 12 V or greater assigned 100\% SOC.}
\label{fig:4}
\end{figure}

The remainder of this section is organized as follows.
Section~\ref{sec:hardware} describes the field-unit hardware and solar
power system, and Section~\ref{sec:site} details the study site and
data collection procedures. Section~\ref{sec:software} presents the
embedded software architecture and fault-tolerance mechanisms, followed
by the two-stage inference algorithm and NGDS calculation in
Section~\ref{sec:algorithm}. Sections~\ref{sec:dataset}
and~\ref{sec:evaluation} describe the dataset preparation,
model-training strategy, and evaluation protocols. Finally,
Sections~\ref{sec:qda} and~\ref{sec:multisite_2026} present the
statistical mapping of NGDS scores onto management practices, the
2026 multi-unit field evaluation, and the remote web dashboard.

\subsection{Hardware Design}
\label{sec:hardware}

\subsubsection{Field Unit and Imaging}

The field unit was mounted on a vineyard trellis post, with the electronics housed in a weatherproof enclosure to protect them from rain and dust during outdoor deployment (Fig.~\ref{fig:fieldunit}). The enclosure was positioned approximately 1 m above the grapevine canopy, with the camera oriented toward the target vine. The solar panel was mounted at the top of the post and oriented toward the south to maximize solar exposure. This fixed mounting arrangement maintained a consistent viewing geometry throughout the growth stage.

The field unit integrated image acquisition, edge processing, power management, and LoRa communication within a single weatherproof enclosure(Fig.~\ref{fig:fieldunit}). 
The NVIDIA Jetson Orin Nano 8 GB development board (A), equipped with a CUDA-capable GPU, served as the main processor for image acquisition and TensorRT-accelerated inference. A 12.8V, 6000 mAh lithium iron phosphate (LiFePO$_4$) battery (B) and solar charge controller (C) managed the power supply, while a Maduino LoRa board (D) supported wireless data transmission. The Sony IMX477 12.3 MP CMOS camera (E) was integrated into the same enclosure and connected to the Jetson through a MIPI CSI-2 ribbon cable for 4K RGB image acquisition. The solar-panel cable (F) supplied input power to the charge controller (C), and a desiccant pack (G) was included to limit moisture accumulation. A custom LoRa gateway (H), comprising a Raspberry Pi (I) connected to a second Maduino LoRa board and an antenna (J), received data from the field units and forwarded them to the cloud server through an Internet connection.

\subsubsection{Solar Power and Battery Monitoring Circuit}

To enable battery monitoring during field deployment, a custom analog voltage-sensing circuit was developed to estimate the battery state of charge (SOC) from its terminal voltage.
A DC Y-splitter routed one branch of the battery output to the Jetson and the other to a resistive voltage divider ($R_1=10$ k$\Omega$, $R_2=2.2$ k$\Omega$; Fig.~\ref{fig:3}), 
which scaled the battery voltage to within the input range of the Maduino's ATmega328 microcontroller. 
The divider values were chosen to keep the source impedance ($\sim$1.8 k$\Omega$) below the 10 k$\Omega$ limit required by the ATmega328's built-in 10-bit ADC for accurate readings. 
The Maduino sampled the divided voltage on analog pin A1 (ten-sample average) and forwarded the result to the Jetson over UART on request. 
The raw ADC value was converted to battery voltage using $V_{\mathrm{bat}} = k\cdot\mathrm{ADC_{raw}}$, where $k = 0.01461$ V\,count$^{-1}$ was determined from paird ADC readings and multimeter measurements (an ADC reading of 739 at a measured 10.8\,V). 
The resulting battery voltage was mapped to an approximate SOC by piecewise-linear interpolation between the calibration points shown in Fig.~\ref{fig:4}.

\subsubsection{Embedded Software and Fault Tolerance}
\label{sec:software}
The runtime software used a modular multiprocess architecture comprising three independently monitored stages: image capture through a GStreamer subprocess, model inference using the TensorRT engine, and LoRa transmission of the resulting severity and status data. Each stage included dedicated error handling. If image capture failed because of a transient camera or GStreamer error, the system repeated the capture attempt once. If the second attempt also failed, the current cycle was skipped and the corresponding error code was recorded. The battery state of charge was read at the beginning of each cycle. When the SOC fell below 40\%, the system suspended image capture and inference and transmitted only a status heartbeat containing the current SOC and power-saving code~05. Full operation resumed when the SOC exceeded 45\%. This hysteresis prevented repeated switching near the cutoff threshold. Each cycle was assigned one of six status codes (Table~\ref{tab:errors}). The selected code was included in the LoRa payload and forwarded to the server to support remote fault identification.

\begin{table}[htbp]
\centering
\caption{Error code system embedded in each 30-byte LoRa payload.}
\label{tab:errors}
\begin{tabular}{c p{0.75\linewidth}}
\toprule
Code & Meaning \\
\midrule
00 & All stages completed successfully \\
01 & Image capture failed (camera/GStreamer error) \\
02 & Model inference failed (TensorRT engine error) \\
03 & Battery voltage read failed (ADC/serial error) \\
04 & LoRa data transmission failed \\
05 & Low-battery power-save \\
\bottomrule
\end{tabular}
\end{table}

The entire pipeline was registered as a Linux \texttt{systemd} service that
starts automatically after  60-second after boot. This delay provided a brief 
window for local intervention through the login console before automated operation began.
If a fatal error caused the Jetson to reboot, the service restarted automatically. Because the Jetson carrier board lacked a battery-backed real-time clock (RTC) and the field unit had no network connection for Network Time Protocol (NTP) synchronization, software-based time persistence was implemented. The current timestamp was written to local storage before each capture cycle and after each sleep transition. During startup, the most recently stored timestamp was used to restore the system clock. Under normal scheduled operation, this procedure maintained a monotonically increasing clock and limited the timestamp error to no more than one 30-min cycle, which was sufficient for duty-cycle control and time-stamping the transmitted records.

\subsection{Study Site and Image Acquisition}
\label{sec:site}
The image dataset used for model development and internal patch-level
evaluation was collected during the 2025 growing season from a mature
\emph{Frontenac} grapevine block at the University of Nebraska--Lincoln
Kimmel Education and Research Center in Nebraska City, Nebraska.
Three D-CAM field units were mounted on trellis posts
within the block. The cameras were positioned approximately 1.52~m
above ground and about 1~m above the target canopy. Following the
scheduled duty cycle, each unit captured one high-resolution RGB canopy
image ($4656 \times 3496$~pixels) every 30~min between 07:00 and 19:00.
Data acquisition was conducted from May 28 to September 21, 2025,
yielding 6,142 raw field images across the three units. The collection
period covered multiple phenological stages, including bud development,
flowering, and fruit set, while capturing natural variation in
illumination, canopy structure, and leaf occlusion.

\begin{figure*}[!t]
\centering
\includegraphics[width=\linewidth]{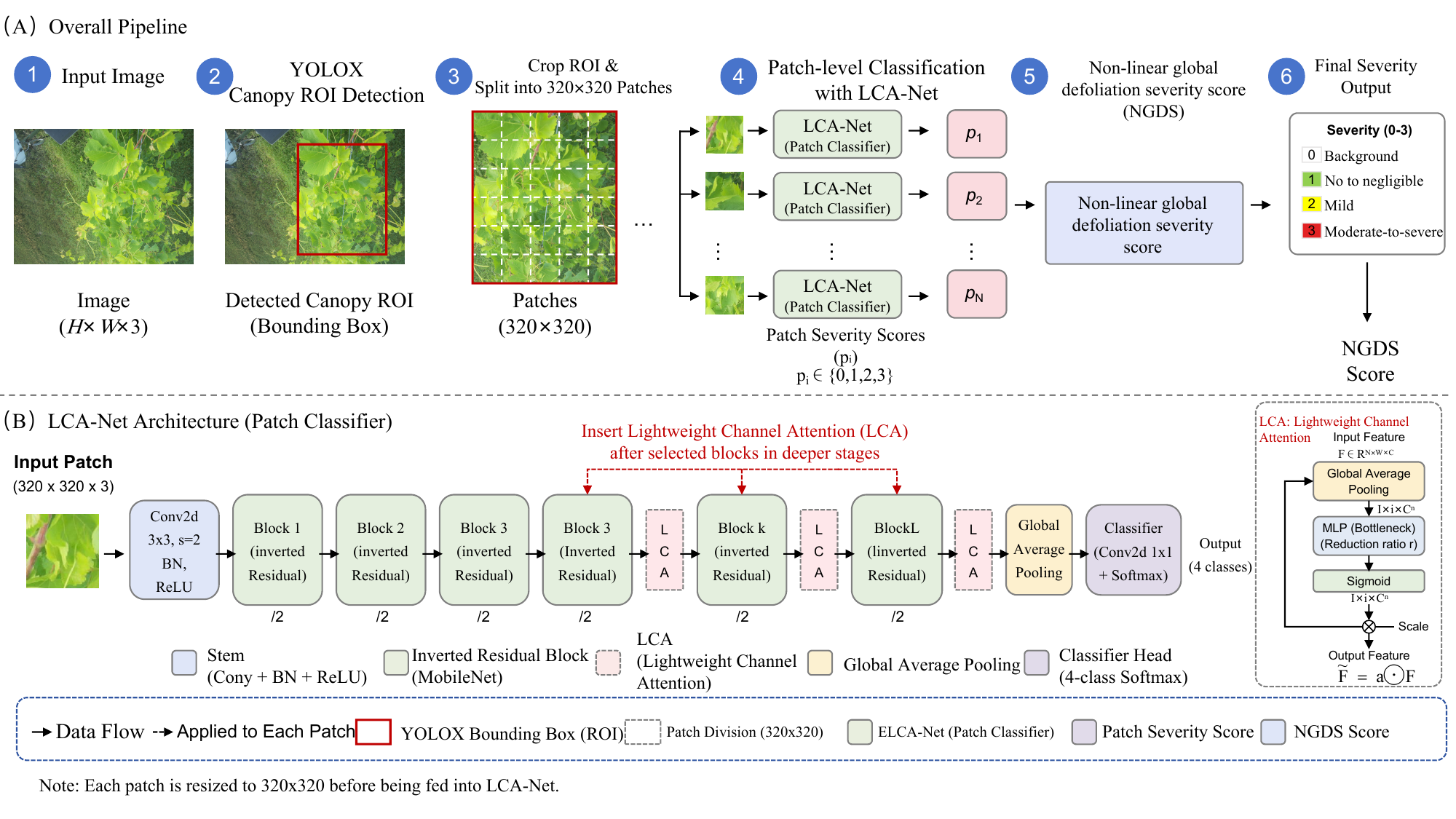}
\caption{(A)~Two-stage pipeline: (1)~input; (2)~YOLOX-Nano canopy detection;
(3)~crop and tile into $320\times320$ patches; (4)~four-class patch
classification; (5)~NGDS aggregation; (6)~severity output. (B)~LCA-enhanced
classifier architecture.}
\label{fig:pipeline}
\end{figure*}

\subsection{Image Annotation and Inference Pipeline}
\label{sec:algorithm}
The workflow of image analysis comprised two stages (Fig.~\ref{fig:pipeline}). In Stage~1, YOLOX-Nano localized the grapevine cordon region within each 4K field image. The detected region was cropped and divided into $320\times320$-pixel patches. In Stage~2, a classifier enhanced with lightweight channel attention (LCA) assigned each patch to the severity classes. The patch-level predictions were then aggregated to calculate the canopy-level Non-linear Global Defoliation Severity (NGDS) score.

\subsubsection{Ground-Truth Annotation and Patch Preparation}

Annotations were prepared separately for the two stages of the pipeline. 
For Stage~1, the target grapevine cordon region was manually delineated with a bounding box in each image using LabelMe. These annotations were used to train the YOLOX-Nano detector. The trained detector was subsequently applied to the field images used for classifier dataset preparation, and each detected cordon region was cropped and divided into $320\times320$-pixel patches. 
Each patch was manually assigned to one of five original patch classes: background, no to negligible damage, mild, moderate, or severe damage. This process yielded 29{,}674 labeled patches for the development and evaluation of the Stage~2 classifier.

\subsubsection{Stage 1: Grapevine Cordon Localization}

YOLOX-Nano~\cite{ge2021}, an anchor-free detector with approximately 0.91 M parameters, was used to localize the grapevine cordon region in each field image. 
The detector operated at an input resolution of $416\times416$ pixels. The selected bounding box was mapped to the coordinates of the original 4K image and used to crop the cordon region for subsequent patch classification. 
Two quality-control conditions were applied before Stage~2. If no valid cordon region was detected, or if the detected bounding box did not meet the minimum size of $512\times512$ pixels in the original image, patch classification was not performed and the cycle was recorded as an invalid observation.

\subsubsection{Stage 2: LCA-Enhanced Four-Class Patch Classification}

The patch dataset was originally annotated using five classes: background, no to negligible damage, mild damage, moderate damage, and severe damage
(Table~\ref{tab:dataset}). The moderate and severe classes contained 1{,}134 and 320 patches, respectively, representing only 4.9\% of the complete dataset.
During model development, these two classes were combined into a single damaged class, resulting in four classes: background, no to negligible damage, mild damage, and moderate-to-severe damage. 
The merged damaged class contained 1{,}454 patches, reducing the ratio between the largest and smallest classes from approximately 40:1 to 8.8:1. 
The original five-class and consolidated four-class formulations were compared as part of the model evaluation.

\begin{figure*}[!t]
\centering
\includegraphics[width=\linewidth,trim=0 30 0 0,clip]{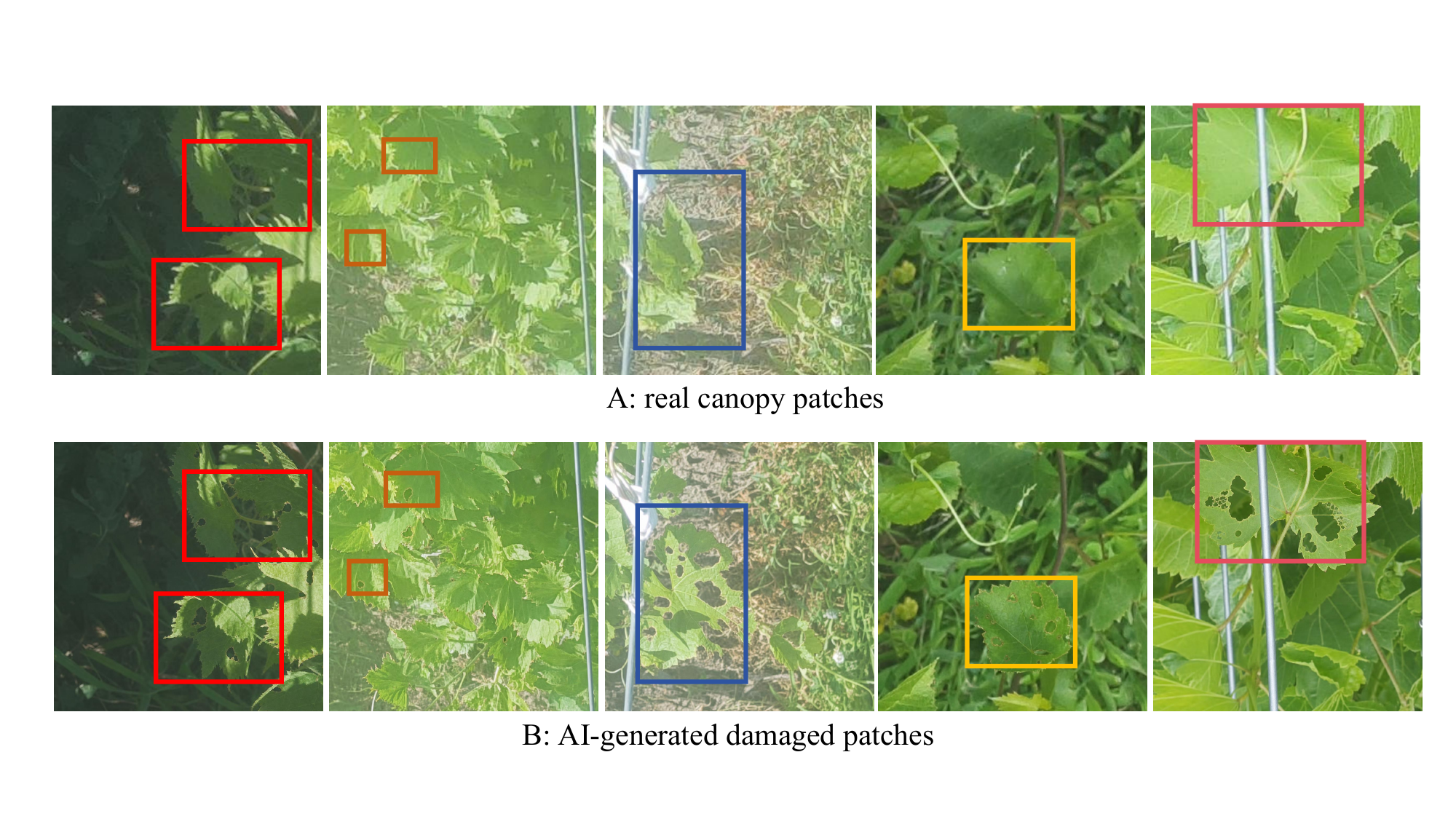}
\caption{Row~(A): real no to negligible damage canopy patches. Row~(B): AI-generated damaged
images with realistic insect-chewing damage (irregular holes, skeletonized
patches, organic edges) via prompt-driven synthesis with randomized hole count
(5--9). The synthetic images preserve natural leaf textures and lighting.}
\label{fig:aiaug}
\end{figure*}

\begin{table}[htbp]
\centering
\caption{Distribution of the 29{,}674 labeled patches among the five original classes. The moderate damage and severe damage classes were subsequently combined into a single moderate-to-severe damage class, reducing the largest-to-smallest class ratio from approximately $\sim$40:1 to 8.8:1.}
\label{tab:dataset}
\setlength{\tabcolsep}{4pt}
\begin{tabular}{c >{\raggedright\arraybackslash}p{2.6cm} r r}
\toprule
Class ID & Label & Count & Proportion (\%) \\
\midrule
0 & Background & 10{,}750 & 36.2 \\
1 & No to negligible & 4{,}736  & 16.0 \\
2 & Mild       & 12{,}734 & 42.9 \\
3 & Moderate   & 1{,}134  & 3.9  \\
4 & Severe     & 320      & 1.1  \\
\midrule
  & Total      & 29{,}674 & 100.0 \\
\bottomrule
\end{tabular}
\end{table}

To increase the representation of the damaged class, approximately 400 synthetic damaged patches were generated from no to negligible damage training patches using the OpenAI \texttt{gpt-image-2} model in image-editing mode at a resolution of $1024\times1024$ pixels. 
For each source patch, one of two prompt templates representing moderate or severe insect-chewing damage was selected at random. 
The prompts specified irregular holes, interveinal skeletonization, nonuniform leaf margins, and a randomized hole count of 5--9 per image. 
All generated patches were assigned to the merged damaged class. Representative real and synthetic patches are shown in Fig.~\ref{fig:aiaug}. 
Synthetic patches were included only in the training partition, and class-balanced sampling was applied with a target of approximately 2{,}000 training samples per class. 

\subsubsection{Non-linear global defoliation severity score (NGDS)}
\label{sec:ngds}

Patch-level predictions were aggregated into a canopy-level Nonlinear Global Defoliation Severity (NGDS) score using an adaptation of the weighted class-frequency structure of the disease severity index~\cite{chiang2017}. 
Let $N_v$ denote the total number of valid non-background patches, including patches classified as no to negligible damage, mild, or moderate-to-severe damage. 
Let $N_{\mathrm{mild}}$ and $N_{\mathrm{mod-to-sev}}$ denote the numbers of patches classified as mild damage ($p_i=2$) and moderate-to-severe damage ($p_i=3$), respectively.
Their corresponding proportions were calculated as $r_{\mathrm{mild}}=N_{\mathrm{mild}}/N_v$ and $r_{\mathrm{mod-to-sev}}=N_{\mathrm{mod-to-sev}}/N_v$. 
A weighted base score was calculated as
\begin{equation}
S_{base} = 0.15 \cdot r_{mild} + 1.0 \cdot r_{mod-to-sev}
\end{equation}
A nonlinear amplification factor was then applied according to the proportion of moderate-to-severe patches:
\begin{equation}
M = 1.0 + 20.0 \cdot (r_{mod-to-sev})^2
\end{equation}
The final NGDS score was calculated as
\begin{equation}
\mathrm{NGDS} = \min(50\%, S_{base} \cdot M \cdot 100\%)
\end{equation}
The base term $S_{\mathrm{base}}$ followed the weighted class-frequency structure of the disease severity index~\cite{chiang2017}, 
while the coefficients were defined specifically for the NGDS formulation. 
A weight of 0.15 was assigned to mild-damage patches to limit their contribution relative to patches classified as moderate-to-severe damage, which received a weight of 1.0. 
The coefficient of 20 controlled the strength of the quadratic amplification term, increasing the sensitivity of NGDS as the proportion of moderate-to-severe patches increased. 

The upper reporting limit was set to 50\% because field images representing defoliation above this level were sparsely represented in the development dataset. 
In addition, at high defoliation levels, completely removed or abscised leaves appear as background, and a single post-damage image does not provide a reliable reference for estimating the original leaf area. 
The NGDS values exceeding 50\% were therefore reported at the upper limit. 
An NGDS value of 50\% consequently indicates severity at or above the supported reporting range and should not be interpreted as an exact measurement of 50\% leaf-area loss.

\subsection{Dataset Partitioning and Model Training}
\label{sec:dataset}
The dataset contained 29{,}674 field-collected patches, which were stratified by class and divided into training (70\%), validation (15\%), and test (15\%) partitions. The split was performed before synthetic augmentation. Synthetic patches were added only to the training partition, whereas the validation and test partitions contained only field-collected patches.

All Stage~2 classifier configurations were trained for 80 epochs using AdamW ($\beta_1=0.9$, $\beta_2=0.999$, and weight decay $=10^{-4}$). The initial learning rate was $10^{-3}$ and was reduced to $10^{-6}$ using a cosine-annealing schedule. Training used a batch size of 16, an input resolution of $320\times320$ pixels, label smoothing of 0.1, and automatic mixed precision. For each configuration, the checkpoint with the highest validation macro-F1 score was retained for testing. Training was performed on two NVIDIA Tesla V100 GPUs, each with 32,GB of memory, at the Holland Computing Center, University of Nebraska--Lincoln. The selected model was then converted to an FP16 TensorRT engine for on-device inference.

\subsection{Model Evaluation Protocol}
\label{sec:evaluation}

Patch-level classification performance was evaluated on the held-out test partition using overall accuracy, macro-F1, weighted-F1, binary healthy-versus-unhealthy accuracy (H-Acc), and damaged-class accuracy (D-Acc). Overall accuracy represented the proportion of all test patches assigned to the correct class. Macro-F1 was calculated as the unweighted mean of the class-specific F1 scores, whereas weighted-F1 accounted for the number of samples in each class. Macro-F1 was emphasized because it gave equal importance to the underrepresented and dominant classes. H-Acc and D-Acc were included to evaluate healthy-versus-unhealthy discrimination and recognition of the moderate-to-severe damage class, respectively.
Two model comparisons were conducted. First, a configuration analysis examined the effects of label-space granularity (five versus four classes), class-balancing strategy, flat versus hierarchical classification, and synthetic-image augmentation. Second, six backbone architectures, including ResNet-18 and ResNet-50, MobileNetV2, MobileNetV3-Small, ShuffleNetV2, and the LCA-enhanced classifier, were evaluated using the same four-class training protocol, with approximately 2{,}000 training samples per class. The architectures were trained and evaluated using the same data partitions and training settings to provide a consistent basis for comparison.

The effect of patch overlap was evaluated using overlap ratios of 0\%, 15\%, 25\%, and 50\%, corresponding to strides of 320, 272, 240, and 160 pixels, respectively, for the $320\times320$-pixel patches. When multiple patches covered the same image region, their predictions were combined by majority vote. Ties were resolved in favor of the class with the highest mean softmax probability. On-device runtime was measured on the Jetson Orin Nano operating in the 15,W power mode using the TensorRT FP16 engine. Long-term system operation was evaluated separately during field deployment.

\subsection{NGDS Management Thresholds}
\label{sec:qda}
Image-level NGDS agreement was evaluated using 86 field-collected canopy images, comprising 70 images from the 2025 dataset and 16 images collected during the 2026 field deployment. Because the 2025 dataset contained few examples of action-level damage, the 2026 images were selected to increase representation of this category and provide a broader severity range for assessment by the entomologist. These 86 images were independent of the images used to generate the 29{,}674-patch dataset, and neither the images nor their derived patches were included in classifier training, validation, or patch-level testing. For each image, the detected cordon region was divided into non-overlapping $320\times320$-pixel patches using the same procedure applied during inference.

An entomologist reviewed each image without access to the model output and assigned it to one of three canopy-level categories: \textbf{(1) No to negligible damage}, requiring no action; \textbf{(2) Mild damage}, indicating that continued monitoring was warranted but no immediate treatment was needed; or \textbf{(3) Action-level damage}, indicating that prompt on site scouting and consideration of insecticide treatment were needed. The absolute difference between the mean model-derived and ground-truth NGDS values and the root mean square error (RMSE) were evaluated in each canopy-level catgories. The former quantified group-level deviation, whereas RMSE summarized image-level error. Both metrics were calculated overall and by expert-assigned category.
For image $i$, $\hat{s}_i$ represents the model-derived NGDS score and $s_i^{\mathrm{GT}}$ represents the ground-truth NGDS score. For an evaluated group containing $N$ images, the absolute mean difference and RMSE were calculated as follows:
\begin{align}
|\Delta\text{GT}| &= \left| \frac{1}{N}\sum_{i=1}^{N}\hat{s}_i - \frac{1}{N}\sum_{i=1}^{N}s_i^{\text{GT}} \right|, \label{eq:dgt}\\
\text{RMSE} &= \sqrt{\frac{1}{N}\sum_{i=1}^{N}\bigl(\hat{s}_i - s_i^{\text{GT}}\bigr)^{2}}. \label{eq:rmse}
\end{align}

Quadratic discriminant analysis (QDA) was fitted to the paired model-derived NGDS scores and expert-assigned categories~\cite{hastie2009elements}. QDA was selected because it allows the NGDS distribution within each category to have a different variance. Classification performance was evaluated using leave-one-out cross-validation because of the limited size of the image-level evaluation set. A final QDA model was then fitted to all 86 paired observations to estimate two decision boundaries separating the three expert-assigned categories. The resulting boundaries were used in the subsequent categorical analyses.

\subsection{2026 Multi-Unit Field Evaluation and Remote Monitoring}
\label{sec:multisite_2026}

A field evaluation was conducted from August~12 to September~16, 2026, at the Kimmel Education and Research Center. Two D-CAM units were installed in different grapevine locations within the research center. Unit~A monitored a young grapevine planting, whereas Unit~B monitored a mature grapevine block. The units experienced similar weather conditions but imaged grapevines that differed in age, canopy density, and structure. Both units used the same hardware configuration, duty cycle, image-processing procedure, and inference pipeline. Data collected during the 2026 deployment were reserved for field evaluation and were not used for model development or NGDS threshold derivation.

During each operating cycle, the field units transmitted the NGDS score, battery state of charge, and device status through the custom LoRa gateway to the cloud server. The server stored the time-stamped records in a MySQL database and displayed unit-specific time series of NGDS and battery state of charge, together with the current device status, on the web dashboard. The dashboard enabled remote monitoring of canopy defoliation, system status, and available battery power throughout the deployment.

LoRa communication reliability was also evaluated during the 2026 field deployment by calculating the packet loss rate (PLR). For each field unit, packets recorded in the local transmission log were matched by device ID and timestamp to the corresponding packets received by the gateway. PLR was calculated as:

\begin{equation}
\label{eq:plr}
    \mathrm{PLR} = \frac{\mathrm{Total\ Packets\ Sent} - \mathrm{Total\ Packets\ Received}}{\mathrm{Total\ Packets\ Sent}} \times 100\%.
\end{equation}

\section{Results and Discussion}
\subsection{Two-Stage Model Performance}
\subsubsection{Stage 1: Grapevine Cordon Localization}
The YOLOX-Nano detector was evaluated on the held-out test set of 1500 canopy images. The detector achieved an mAP@0.5 of 93.2\%, with a precision of 94.1\% and a recall of 92.5\%. The mean inference time was 39~ms per image on the Jetson Orin Nano using the TensorRT FP16 engine. These results supported the use of YOLOX-Nano for extracting the cordon region before patch-level severity classification.

\subsubsection{Stage 2: Patch-Classifier Configuration Analysis}

\begin{table*}[tp]
\centering
\footnotesize
\setlength{\tabcolsep}{3.5pt}
\caption{Test-set performance of the Stage~2 patch-classifier configurations. H-Acc denotes binary healthy-versus-unhealthy accuracy, and D-Acc denotes moderate-to-severe damage-class accuracy.}
\label{tab:ablation}

\begin{tabular}{@{}r p{4.3cm} c c c c c c c@{}}
\toprule
ID & Classifier configuration   & Classes & \shortstack{Target patches/class} & \shortstack{AI augmentation} & \shortstack{Accuracy (\%)} & Macro-F1 & H-Acc & D-Acc \\
\midrule
1  & Decoupled classifier + focal loss & 5 & 1,500 & No  & 78.4 & 0.643 & 0.766 & 0.452 \\
2  & Decoupled classifier + focal loss & 5 & 3,000 & No  & 78.0 & 0.657 & 0.759 & 0.482 \\
3  & Decoupled classifier + focal loss & 5 & 5,000 & No  & 82.6 & 0.685 & 0.811 & 0.530 \\
4  & Hierarchical classifier (two models) & 5 & --- & No & 73.9 & 0.596 & 0.699 & 0.398 \\
\midrule
5  & LCA-enhanced classifier & 4 & 2,000 & No & 80.3 & 0.742 & 0.789 & 0.617 \\
6  & LCA-enhanced classifier + validation-based balancing & 4 & 2,000 & No & 77.3 & 0.706 & 0.761 & 0.534 \\
7  & LCA-enhanced classifier & 4 & Natural distribution & No & \textbf{83.6} & 0.768 & 0.822 & 0.617 \\
\textbf{8} & \textbf{LCA-enhanced classifier (deployed model)} & \textbf{4} & \textbf{2,000} & \textbf{Yes} & \textbf{83.6} & \textbf{0.838} & \textbf{0.835} & \textbf{0.830} \\
9  & LCA-enhanced classifier & 4 & 5,000 & Yes & 81.6 & 0.763 & 0.817 & 0.704 \\
\midrule
10 & ResNet-50 classifier & 4 & 2,000 & Yes & 79.1 & 0.746 & 0.803 & 0.781 \\
\bottomrule
\end{tabular}%

\end{table*}

The evaluated Stage~2 configurations produced macro-F1 scores ranging from 0.596 to 0.838, indicating substantial differences among the class formulations, balancing strategies, and training configurations (Table~\ref{tab:ablation}). 

\textbf{Effect of class resampling in the five-class setting (rows~1--3).}
ncreasing the per-class resampling target from 1{,}500 to 5{,}000 patches increased macro-F1 from 0.643 to 0.685 and 
damaged-class accuracy from 0.452 to 0.530.  Overall accuracy did not increase consistently with the resampling target, 
decreasing slightly from 78.4\% to 78.0\% at the 3,000-patch target before reaching 82.6\% at the 5,000-patch target. 
Although the 5{,}000-patch configuration performed best among the five-class balancing strategies, its damaged-class accuracy remained relatively low.

\textbf{Flat and hierarchical classification (row~4).} The hierarchical two-model configuration, 
which first separated canopy patches from background and then classified damage severity, produced the lowest overall accuracy (73.9\%), macro-F1 (0.596), and damaged-class accuracy (0.398) among the evaluated configurations. 
Because the second stage only ever sees the first stage's output, misclassifications
propagate and compound rather than being corrected, confirming that a single flat
classifier is preferable for this task.

\textbf{Five- and four-class configurations (rows~3 and~7).} Consolidating the two
most under-represented severity levels (moderate and severe) into a single
\emph{moderate to severe} class raises macro F1 from 0.685 to 0.768 ($+8.3$ points) and increased
moderate to severe damaged-class accuracy from 0.530 to 0.617. However, the two configurations also differed in classifier design, loss function, 
and class-balancing strategy. Therefore, the observed performance difference reflects the overall four-class LCA configuration and cannot be attributed solely to merging the moderate and severe classes.

\textbf{Effect of class resampling in the four-class setting (rows~5--7).} The model trained without class resampling achieved a macro-F1 of 0.768, compared with 0.742 for resampling to 2{,}000 samples per class and 0.706 for validation-based resampling. Overall accuracy showed the same pattern and was highest for the model trained without resampling (83.6\%). 
Thus, neither of the evaluated resampling strategies improved classification performance under the four-class formulation.

\textbf{Effect of augmentation with AI-generated patches (rows 5--8).} 
Rows~5 and~8 used the same four-class LCA-enhanced architecture and a resampling target of 2{,}000 samples per class, 
allowing the effect of synthetic augmentation to be evaluated directly. 
Adding synthetic patches depicting moderate-to-severe damage increased overall accuracy from 80.3\% to 83.6\%, macro-F1 from 0.742 to 0.838, and moderate to severe damaged-class accuracy from 0.617 to 0.830. Representative field-collected and synthetic patches are shown in Fig.~\ref{fig:aiaug}. Among the AI augmented configurations, increasing the resampling target from 2{,}000 to 5{,}000 samples per class reduced macro-F1 from 0.838 to 0.763 and damaged-class accuracy from 0.830 to 0.704 (rows~8 and~9). Based on its macro-F1 and damaged-class accuracy, the four-class LCA model trained with AI-generated patches and a resampling target of 2{,}000 samples per class was selected for D-CAM field deployment.

\textbf{Comparison with ResNet-50 (rows~8 and~10).} 
The 1.08,M-parameter LCA-enhanced classifier achieved higher 
test-set performance than the 23.5,M-parameter ResNet-50 
when both models were trained using the same dataset and training protocol. 
Macro-F1 was 0.838 for the lightweight classifier and 0.746 for ResNet-50, while damaged-class accuracy was 0.830 and 0.781, respectively. 
Although ResNet-50 contained approximately 22 times as many parameters, 
its additional capacity did not improve either primary performance metric under the evaluated conditions. 
The lightweight classifier therefore provided higher classification performance and model size for canopy-level defoliation assessment. 
Its compact architecture reduced model-storage requirements and was well suited to repeated on-device inference in the solar-powered field unit and may also facilitate deployment on other edge platforms with limited memory and computing capacity.

\subsection{Backbone Comparison}
Six Stage~2 classifier backbones were trained using the fixed protocol and evaluated on the same held-out, field-collected patch test set. The evaluated architectures included ResNet-18 and ResNet-50, MobileNetV2, MobileNetV3-Small, ShuffleNetV2, and the LCA-enhanced classifier (Table~\ref{tab:backbones}); the architecture reference for each backbone is listed in the last column.

\begin{table*}[t]
\centering
\caption{Stage~2 patch-classification performance of six backbones trained under the same four-class protocol. The training set was resampled to a target of 2{,}000 samples per class, with AI-generated patches included in the moderate-to-severe damage class. All models were evaluated on the same held-out, field-collected test set. }
\label{tab:backbones}
\begin{tabular}{@{}lcccc l@{}}
\toprule
Backbone & Params (M) & Acc (\%) & F1 (macro) & F1 (weighted) & {Reference} \\
\midrule
ResNet-50            & 23.52 & 79.15 & 0.7462 & 0.7992 & {\cite{he2016resnet}} \\
ResNet-18            & 11.18 & 80.24 & 0.7469 & 0.8095 & {\cite{he2016resnet}} \\
MobileNetV2          & 2.23  & 82.46 & 0.7745 & 0.8291 & {\cite{sandler2018mobilenetv2}} \\
MobileNetV3-Small    & 1.52  & 82.96 & 0.7705 & 0.8337 & {\cite{howard2019}} \\
ShuffleNetV2         & 1.26  & 81.78 & 0.7673 & 0.8231 & {\cite{ma2018shufflenetv2}} \\
LCA-enhanced (ours)  & \textbf{1.08} & \textbf{83.60} & \textbf{0.8383} & \textbf{0.8341} & {-} \\
\bottomrule
\end{tabular}
\end{table*}

\begin{table*}[htbp]
\centering
\caption{Per-category agreement between the predicted NGDS and the expert-derived reference NGDS (GT). Configuration~C is the five-class model; the raw input and the four-class configurations (A, B, D--F) are also shown as panels A--E in Fig.~\ref{fig:NGDScomp}. Category
definitions: Cat~1 = No to negligible damage ($\text{GT} < 9\%$, $N=23$); Cat~2 = Mild
($9\% \le \text{GT} \le 29\%$, $N=51$); Cat~3 = Needs action ($\text{GT} > 29\%$, $N=12$).
Pred = mean predicted NGDS (\%); $|\Delta|$ = $|\text{Pred}-\text{GT}|$ in
percentage points; RMSE per Eq.~\eqref{eq:rmse}, in percentage points.
Best per category in bold.}

\label{tab:NGDS}
\setlength{\tabcolsep}{5pt}
\renewcommand{\arraystretch}{1.15}
\begin{tabular}{cl ccc ccc ccc}
\toprule
     &                                & \multicolumn{3}{c}{\shortstack{\textbf{Cat 1: No to}\\\textbf{negligible damage}}}      & \multicolumn{3}{c}{\textbf{Cat 2: Mild}}         & \multicolumn{3}{c}{\textbf{Cat 3: Needs action}}   \\
     &                                & \multicolumn{3}{c}{\footnotesize Avg GT = 7.4\%} & \multicolumn{3}{c}{\footnotesize Avg GT = 17.0\%} & \multicolumn{3}{c}{\footnotesize Avg GT = 47.0\%} \\
\cmidrule(lr){3-5} \cmidrule(lr){6-8} \cmidrule(lr){9-11}
\textbf{Panel} & \textbf{Configuration}     & Pred            & $|\Delta|$        & RMSE            & Pred             & $|\Delta|$        & RMSE            & Pred             & $|\Delta|$        & RMSE            \\
\midrule
A    &  ground truth            & 7.4             & ---               & ---             & 17.0             & ---               & ---             & 47.0             & ---               & ---             \\
\textbf{B} & \textbf{4-cls, 0\% overlap (our deployed model)} & \textbf{6.6} & 0.8              & \textbf{2.4}    & \textbf{16.9}    & \textbf{0.1}      & \textbf{3.5}    & \textbf{46.5}    & \textbf{0.5}      & \textbf{4.4}    \\
C    & 5-cls, 0\% overlap             & 12.0   & 4.6     & 8.0   & 30.0  & 13.0    & 14.0   & 58.0    & 11.0   & 15.0   \\
D    & 4-cls, 15\% overlap            & 5.9             &      1.5     & 2.8             & 14.0             & 3.0              & 4.3             & 43.0             & 4.0              & 8.1             \\
E    & 4-cls, 25\% overlap            & 6.2             & 1.2              & 2.7             & 13.5             & 3.5              & 5.0             & 41.6             & 5.4              & 8.2             \\
F    & 4-cls, 50\% overlap            & 8.0             & 0.6              & 3.2             & 13.9             & 3.1              & 4.3             & 40.8             & 6.2              & 10.7            \\
\bottomrule

\end{tabular}
\end{table*}
Among the six evaluated backbones, a larger parameter count did not yield better classification performance. 
ResNet-50 and ResNet-18 produced the two lowest values for accuracy, macro-F1, and weighted-F1, despite having 23.52 and 11.18 million parameters, respectively. 
In contrast, all four models with fewer than 2.5 million parameters outperformed both ResNet backbones across the three metrics. The 1.08-million-parameter LCA-enhanced classifier achieved the highest accuracy (83.60\%), macro-F1 (0.8383), and weighted-F1 (0.8341), while also having the lowest parameter count.

Relative to ResNet-50, the LCA-enhanced classifier increased macro-F1 by 0.0921 while using 95.4\% fewer parameters.
 Differences from the strongest compact alternatives were smaller for accuracy and weighted-F1. 
 Accuracy was 0.64 percentage points higher than that of MobileNetV3-Small, whereas weighted-F1 differed by only 0.0004. 
 However, macro-F1 exceeded that of the next-best model, MobileNetV2, by 0.0638.
 Because macro-F1 gives equal weight to each class, this result indicates stronger average performance across classes, although the aggregate metric does not identify which class contributed most to the improvement. The combination of the highest test-set performance and the lowest parameter count supported the selection of the LCA-enhanced classifier for repeated inference on the solar-powered field unit.

\subsection{Effect of Tiling Overlap on NGDS Agreement}
\label{sec:expert_validation}

Image-level agreement between the model-derived and entomologist-derived NGDS values was evaluated under 4-class and 5-class configurations and compared at overlap ratios of 0\%, 15\%, 25\%, and 50\%. For each image, the entomologist-derived ground-truth (GT) NGDS was calculated from the patch-level classifications. Valid patches were classified as no to negligible damage, mild, or moderate-to-severe damage, while background patches were excluded. The resulting expert-assigned patch counts were entered into the NGDS formulation described in Section~2.4.4 to calculate the GT NGDS for each image. All model comparisons used the same LCA-enhanced four-class classifier (Table 3, ID=8), only the tiling overlap was changed.

\begin{figure*}[tp]
\centering

\includegraphics[page=1, width=\linewidth,trim=0 0 280 0,clip]{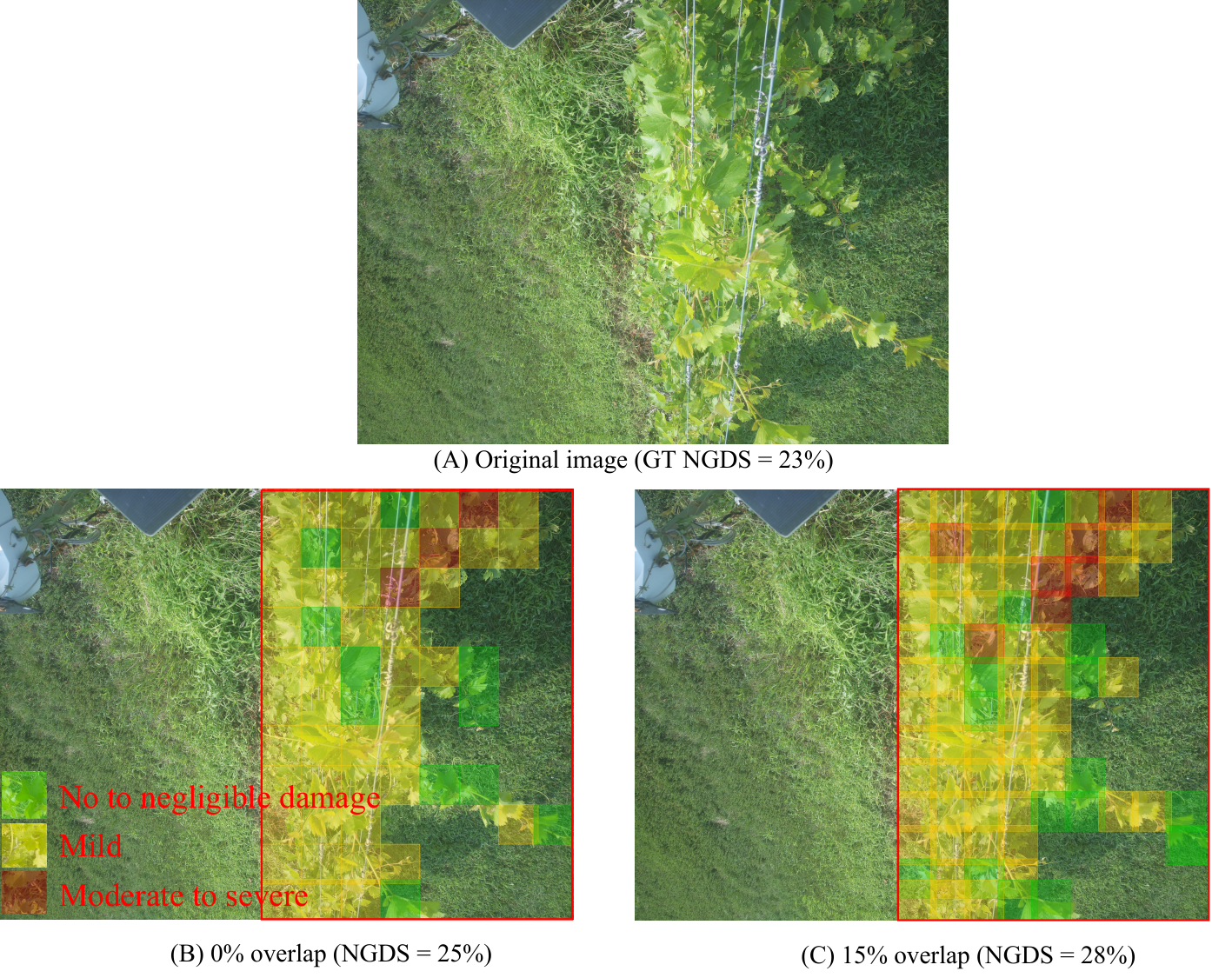}

\vspace{1em}

\includegraphics[page=2, width=\linewidth,trim=0 275 280 0,clip]{./yy.pdf}

\vspace{1em}

\caption{Effect of tiling overlap on patch-level severity predictions for a representative canopy image with an expert-derived ground-truth NGDS of 23\%. (A) Raw image. 
~Raw input. (B)--(E) Severity maps generated by the LCA-enhanced four-class model using overlap ratios of 0\%, 15\%, 25\%, and 50\%, respectively. The corresponding model-derived NGDS values were 25\%, 28\%, 29\%, and 29\%. }
\label{fig:NGDScomp}
\end{figure*}
Under non-overlapping tiling, the four-class configuration showed closer agreement with the expert-derived GT NGDS across all three management categories (Table~\ref{tab:NGDS}). For no to negligible damage, mild-damage, and action-level-damage canopies, the four-class model produced mean NGDS values of 6.6\%, 16.9\%, and 46.5\%, respectively, compared with the corresponding entomologist-derived values of 7.4\%, 17.0\%, and 47.0\%. In contrast, the five-class configuration produced higher mean NGDS values of 12.0\%, 30.0\%, and 58.0\% (Table~\ref{tab:NGDS}). The RMSE also increased from 2.4\%, 3.5\%, and 4.4\% with the four-class configuration to 8.0\%, 14.0\%, and 15.0\% with the five-class configuration. Retaining separate moderate- and severe-damage classes therefore did not improve canopy-level NGDS estimation and resulted in consistently higher severity estimates. The four-class configuration was consequently used for the subsequent tiling-overlap analysis.

Within the four-class configuration, non-overlapping tiling produced the lowest RMSE across the three entomologist-assigned management categories. From 0\% to 50\% overlap, RMSE increased from 2.4 to 3.2 percentage points for no to negligible damage canopies and from 3.5 to 4.3 percentage points for mild-damage canopies (Table~\ref{tab:NGDS}). The effect was more pronounced for action-level damage, for which RMSE increased from 4.4 to 10.7 percentage points (Table~\ref{tab:NGDS}). Increasing the overlap therefore did not improve agreement with the entomologist-derived GT NGDS.

The original canopy image had an expert-derived GT NGDS of 23\% (Fig.~\ref{fig:NGDScomp}A). The non-overlapping configuration produced an NGDS of 25\% (Fig.~\ref{fig:NGDScomp}B). Increasing the overlap to 15\%, 25\%, and 50\% increased the estimated NGDS to 28\%, 29\%, and 29\%, respectively (Fig.~\ref{fig:NGDScomp}C--E). Compared with the non-overlapping result, the overlapping severity maps provided progressively denser coverage of the canopy, and the localized damaged region in the upper canopy was represented by a greater number of patch predictions. This repeated representation of damaged areas was associated with the higher NGDS estimates. Because tiling overlap increased the number of patch inferences and processing time without improving agreement with the expert-derived NGDS, the four-class model with 0\% overlap was selected as the deployed inference configuration.

\subsection{NGDS Distribution Across Management Categories}
\label{sec:ngds_categories}

The management-category analysis was also based on the 86 canopy image test set evaluated by an entomologist. For each image, the developed four-class LCA-enhanced model generated patch-level predictions, which were aggregated to calculate the model-derived NGDS. Independently of the model output, the entomologist assigned each image to one of three management categories: No to negligible damage, Mild damage, or Action-level damage. Quadratic discriminant analysis (QDA) identified NGDS boundaries of 9.32\% and 29.12\%, which were rounded to 9\% and 29\% for management interpretation. The resulting NGDS categories were defined as \textbf{(1)~No to negligible damage},
NGDS~$<$~9\%; \textbf{(2)~Mild damage},
9\%~$\le$~NGDS~$\le$~29\%; and \textbf{(3)~Action-level damage}
(requiring prompt scouting and consideration of insecticide treatment),
NGDS~$>$~29\%. The upper boundary was consistent with the approximately 25\%--30\% action range previously reported for Japanese beetle damage in bunch grapes~\cite{ebbenga2022,guedot2019}. These thresholds mark the operational transitions at which visible foliar damage begins to warrant monitoring (9\%) and at which
intervention becomes economically justified (29\%).

\begin{figure}[htbp]
    \centering
    \includegraphics[width=\linewidth,trim=100 20 100 30,clip]{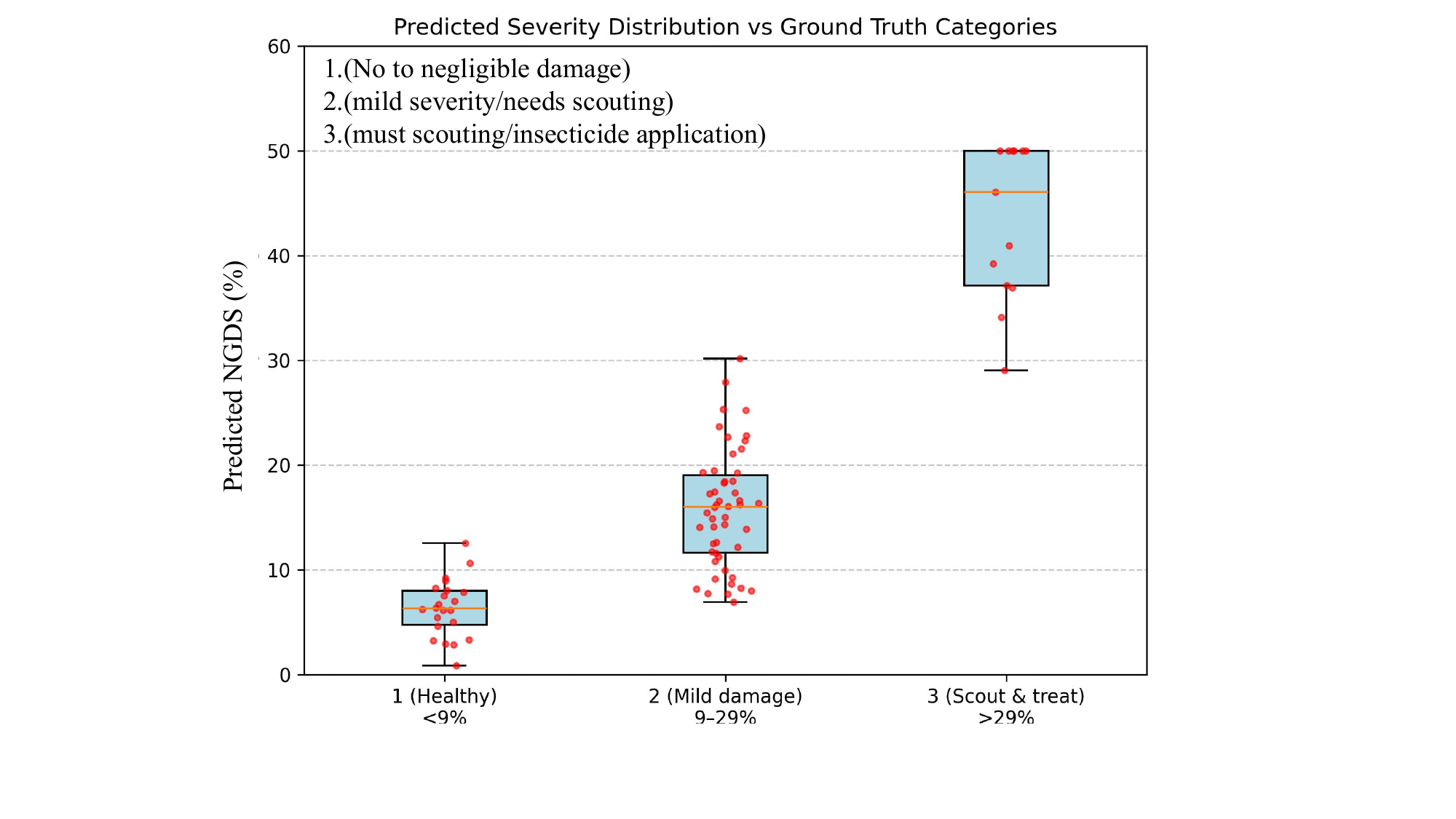}
    \caption{Distribution of model-derived NGDS values for 86 field images grouped by the entomologist-assigned management category. The QDA-derived decision boundaries define \textbf{(1)~No to negligible damage}
    (NGDS~$<$~9\%), \textbf{(2)~Mild damage} (9--29\%), and
    \textbf{(3)~Action-level damage} ($>$~29\%).}
    \label{fig:NGDS_dist}
\end{figure}

The model-derived NGDS distributions were consistent with the entomologist-assigned management categories (Figure~\ref{fig:NGDS_dist}). Median NGDS values were 6.2\%, 16.1\%, and 46.1\% for No to negligible damage, Mild damage, and Action-level damage, respectively. The interquartile range of each distribution remained within its corresponding QDA-derived NGDS band, although some individual observations near the decision boundaries overlapped adjacent categories.

\begin{figure*}[!t]
\centering

\includegraphics[page=1, width=\linewidth,trim=0 70 0 80,clip]{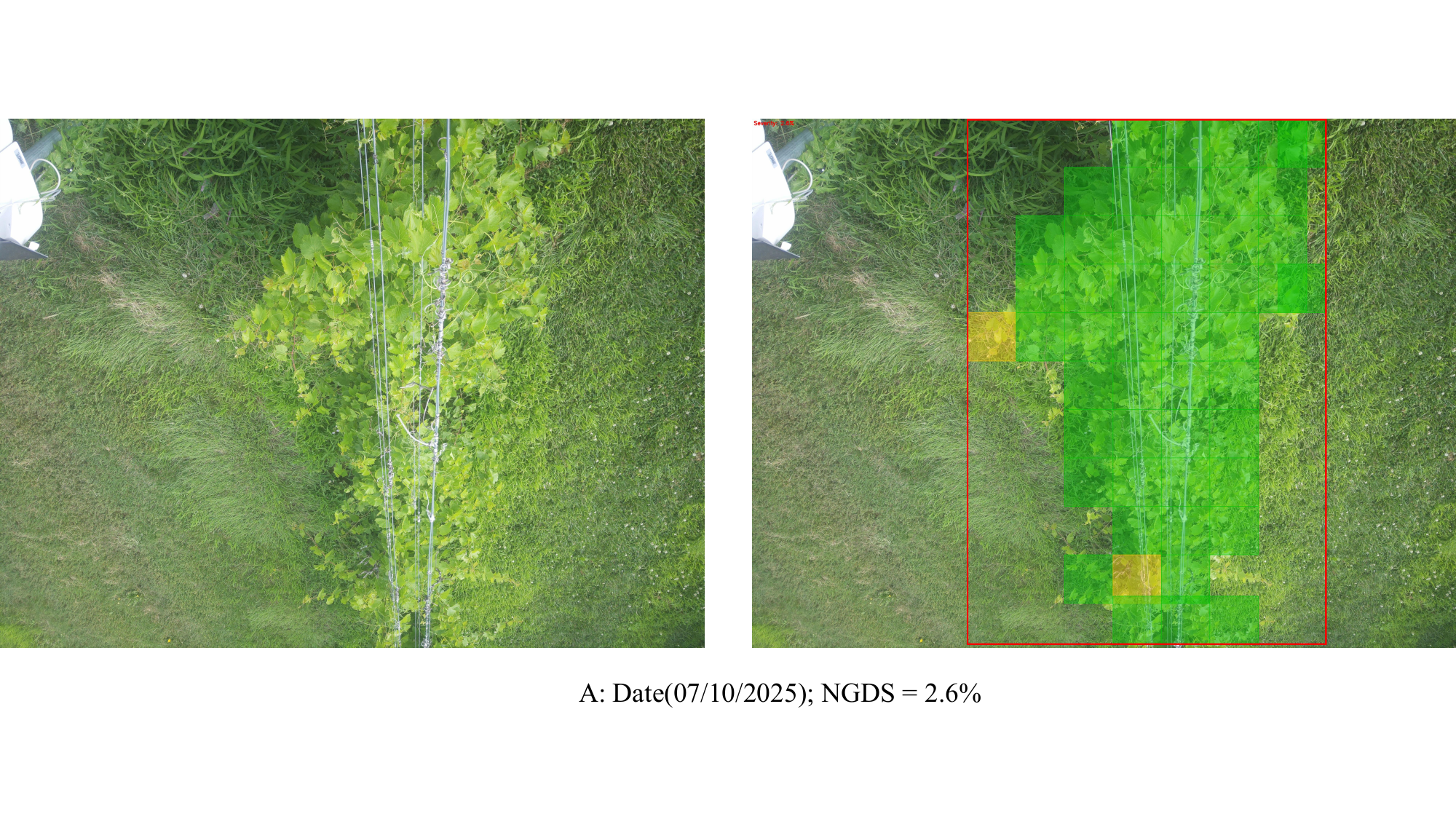}

\vspace{1em}

\includegraphics[page=2, width=\linewidth,trim=0 70 0 80,clip]{./y.pdf}

\vspace{1em}

\includegraphics[page=3, width=\linewidth,trim=0 70 0 80,clip]{./y.pdf}

\caption{Visualization across three NGDS categories: no to negligible damage (A), mild
(B), and action level damage (C).}
\label{fig:qual}
\end{figure*}

The NGDS formulation assigns a lower weight to patches classified as mild damage and increases the aggregate score nonlinearly as the proportion of moderate-to-severe-damage patches increases. This weighting was intended to limit the contribution of minor foliar injury while retaining sensitivity to increasing canopy damage. In the 86-image evaluation set, the resulting scores increased across the entomologist-assigned management categories, and the QDA-derived thresholds provided a basis for interpreting the continuous NGDS. 

Representative on-device inference outputs acquired in mid-July, mid-August, and mid-September illustrate the seasonal increase in canopy defoliation, which was primarily associated with Japanese beetle feeding at the study site (Fig.~\ref{fig:qual}). Each example pairs the original field image on the left with the corresponding patch-classification output on the right. The mid-July No to negligible damage example (A) was dominated by patches classified as no to negligible damage. By mid-August, the Mild-damage example (B) contained a greater proportion of mild and moderate-to-severe damage patches. In mid-September, moderate-to-severe damage patches were distributed throughout the canopy in the Action-level-damage example (C). Non-canopy patches were classified as background and excluded from the NGDS calculation. The corresponding NGDS values were 2.6\%, 23\%, and 48\% for panels A--C, respectively. 

\subsection{Runtime Performance}
 
The deployed pipeline generated a canopy-level NGDS from each 4K image in approximately 1.5~s on the Jetson Orin Nano using TensorRT FP16 inference. Canopy detection required approximately 39~ms, whereas classification of all patches within the detected canopy region required approximately 1,505~ms and accounted for most of the inference latency. Model inference represented less than 0.1\% of the 30-min sampling interval. The complete operating cycle, measured from system wake-up through image acquisition, inference, and repeated LoRa transmissions, lasted approximately 57--60,s, representing 3.2--3.3\% of the sampling interval. The unit then returned to the low-power suspend-to-RAM state until the next scheduled cycle.

\begin{figure*}[tp]
\centering
\includegraphics[width=\linewidth,trim=0 0 0 0,clip]{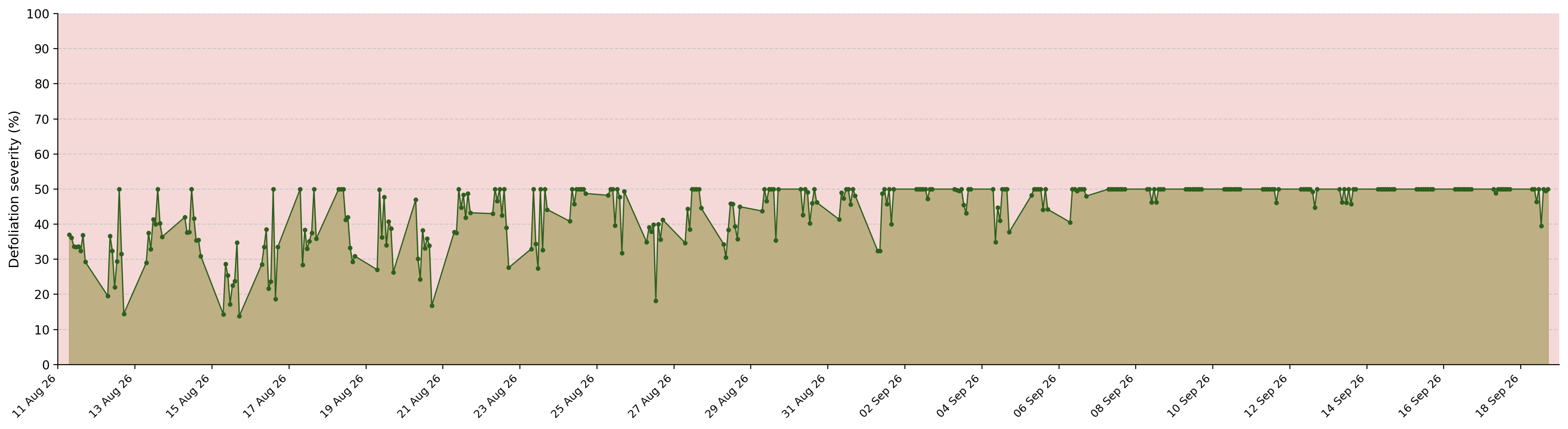}
\caption{Time series of model-derived NGDS values recorded by Unit~B in the mature grapevine block during the 2026 field evaluation. Each point represents the NGDS generated during one scheduled operating cycle and transmitted to the web dashboard. The trace shows short-term variability and an increasing frequency of values near 50\% later in the evaluation period. Because NGDS is capped at 50\% by its formulation, readings at this value represent the upper limit of the metric rather than estimates exceeding 50\%.}
\label{fig:severity_trace}
\end{figure*}

\begin{figure*}[htbp]
\centering
\includegraphics[width=\linewidth,trim=0 0 0 0,clip]{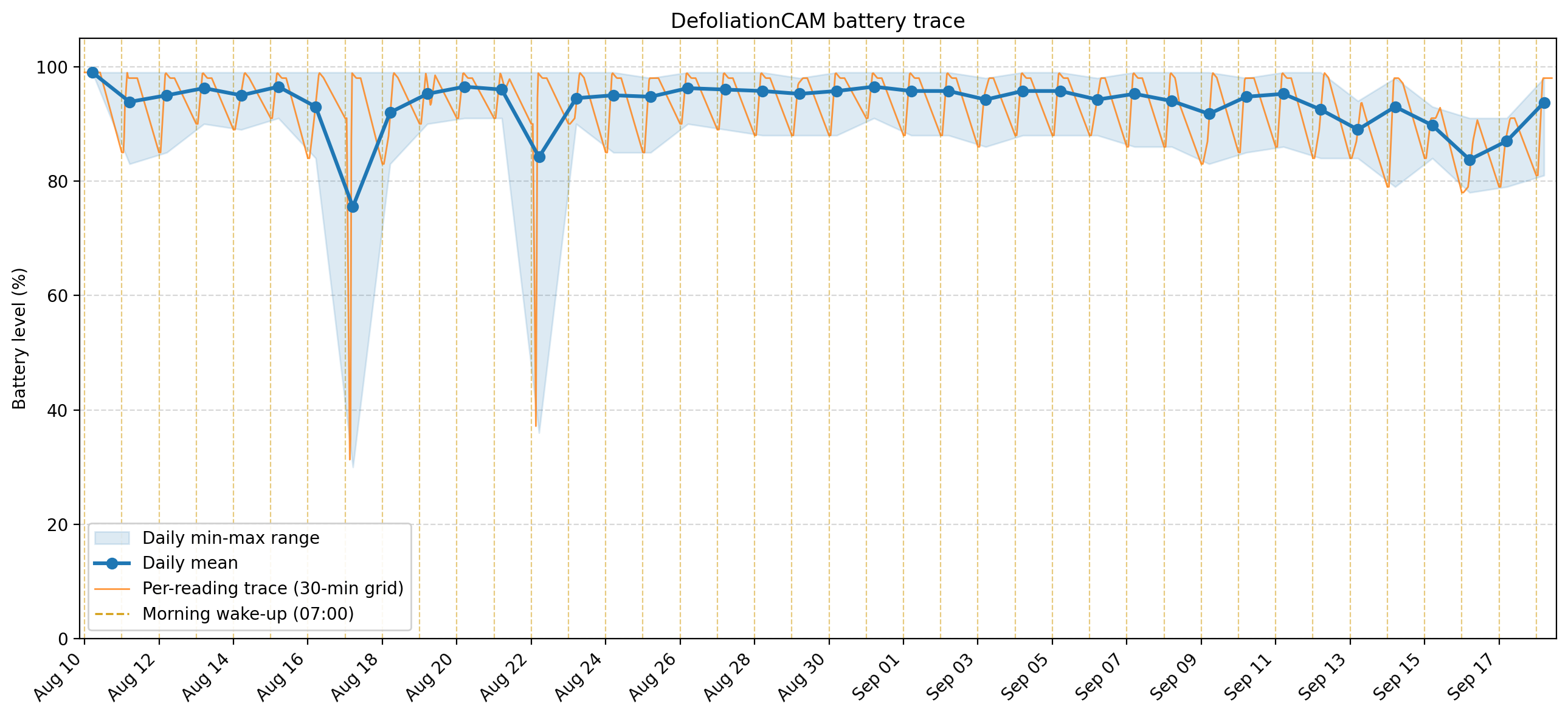}
\caption{Battery state-of-charge record for Unit~B during the 2026 field evaluation. The orange line shows measurements recorded at 30-min intervals during the daily operating window from 07:00 to 19:00. The blue line represents the daily mean, and the shaded band spans the daily minimum and maximum. Gold dashed lines mark the beginning of each daily operating window at 07:00.}
\label{fig:battery_trace}
\end{figure*}

\subsection{2026 Field Deployment and Remote Monitoring}

\begin{figure*}[t!]
\centering
\includegraphics[page=1,width=\linewidth,trim=0 230 0 0]{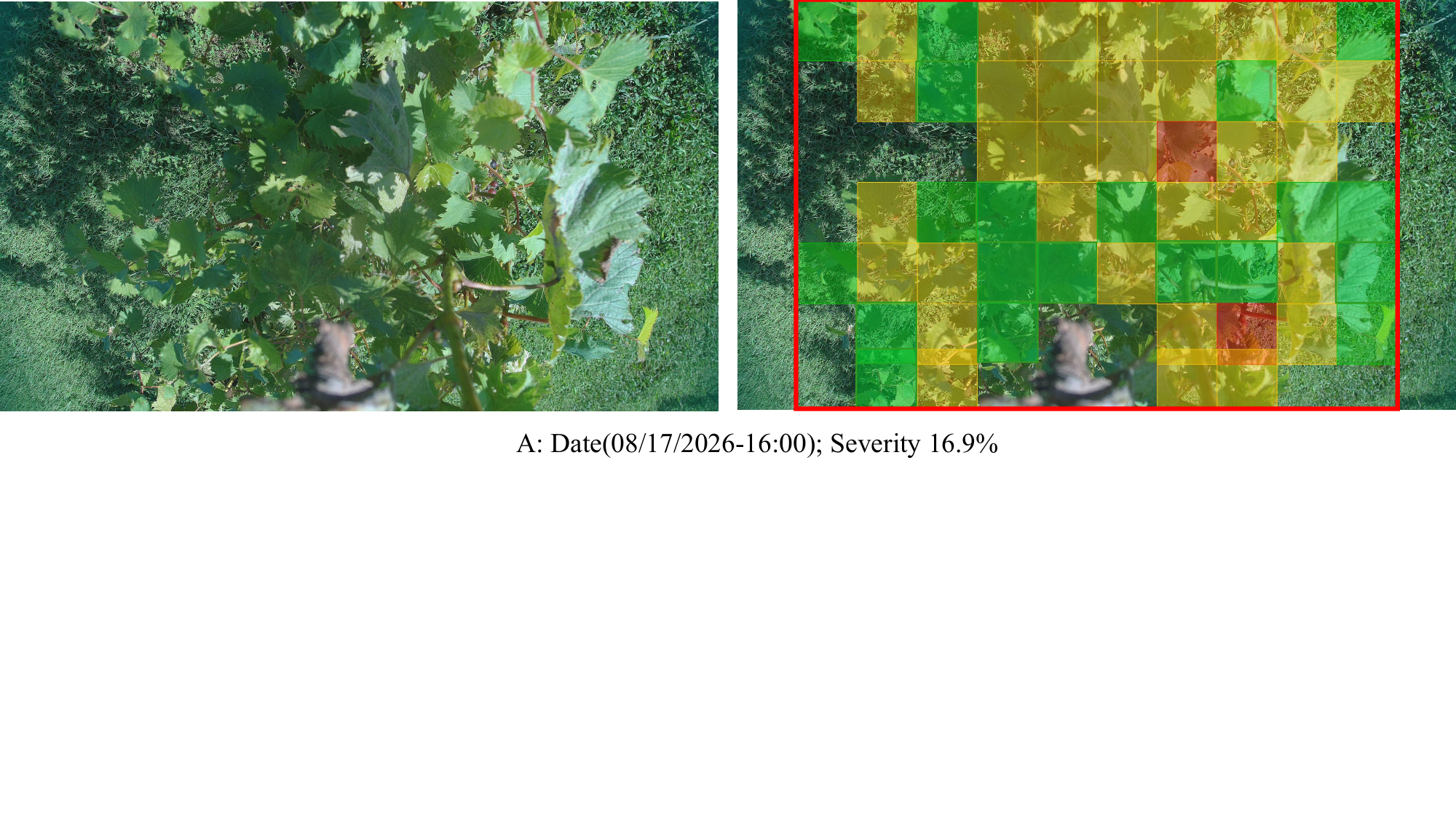}\\[0.4em]
\includegraphics[page=2,width=\linewidth,trim=0 230 0 0]{./u3.pdf}\\[0.4em]
\includegraphics[page=3,width=\linewidth,trim=0 230 0 0]{./u3.pdf}
\caption{Representative Unit~B inference outputs during the 2026 field deployment: (A) August~17, NGDS = 16.9\%; (B) September~1, NGDS = 29.5\%; and (C) September~11, NGDS = 34.4\%. Each example shows the original image (left) and patch predictions (right), with green, yellow, and red representing no to negligible damage, mild, and moderate-to-severe damage, respectively. The increasing NGDS values were consistent with the dashboard trend and observed Japanese beetle feeding damage.}
\label{fig:field_examples}
\end{figure*}

\label{sec:field_deployment}

The 2026 field evaluation was conducted from August~12 to September~16 at the University of Nebraska--Lincoln Kimmel Education and Research Center. Unit~A monitored a young grapevine planting, whereas Unit~B monitored a mature grapevine block. Both units generated time-stamped NGDS, battery state-of-charge (SOC), and device-status records that were transmitted to the web dashboard throughout the deployment. Representative NGDS and battery records from Unit~B are presented in ..~\ref{fig:severity_trace} and~\ref{fig:battery_trace}, respectively.

The NGDS record for Unit~B showed considerable variation among individual sampling cycles, with readings approaching the 50\% upper limit more frequently later in the deployment period. Values at 50\% reflect the cap imposed by the NGDS formulation rather than estimates exceeding this limit (Fig.~\ref{fig:field_examples}). The longer-term increase was consistent with the field-observed accumulation of Japanese beetle feeding damage. Short-term variation among cycles may have reflected changing illumination, leaf movement, and patches near the classification boundaries; therefore, trends over multiple days were more informative than individual readings.

Representative inference outputs from Unit~B further illustrate the temporal change recorded by the dashboard (Fig.~\ref{fig:field_examples}). Images acquired on August~17, September~1, and September~11 produced NGDS values of 16.9\%, 29.5\%, and 34.4\%, respectively. For each example, the left panel shows the original 4K image, and the right panel shows patch-level predictions within the YOLOX-Nano-detected grapevine cordon region outlined in red. Green, yellow, and red indicate patches classified as no to negligible damage, mild damage, and moderate-to-severe damage, respectively. Based on the inference outputs, the progression was consistent with the NGDS time series displayed on the dashboard and with field observations of cumulative Japanese beetle feeding damage.

The battery SOC was recorded at 30-min intervals during the daily operating window from 07:00 to 19:00 (FIg.~\ref{fig:battery_trace}). The first SOC measurement at 07:00 was generally the daily minimum, reflecting the net battery draw during the overnight suspended period. The SOC subsequently increased during daylight hours as solar charging resumed. The daily mean SOC generally remained above 85\%, except for temporary declines around August~17 and 22. The lower SOC observed at the beginning of the daily operating window was likely associated with residual overnight power consumption during suspend-to-RAM and a transient decrease in terminal voltage when the Jetson resumed operation at 07:00. Because SOC was estimated from battery voltage, the morning decline may not represent an equivalent loss of battery capacity. SOC subsequently increased as solar charging resumed.

During the 36-day field evaluation (12 August--16 September 2026), each unit was scheduled to send one data packet---a single NGDS reading---per 30-min capture cycle within the 07:00--19:00 daily operating window, corresponding to approximately 900 scheduled packets per unit. To prevent occasional LoRa link errors from causing data loss, each packet was broadcast five times per cycle, and a packet was counted as received if at least one copy was decoded at the gateway. Even with this redundancy, about two cycles per day were missed per unit: Unit~A and Unit~B successfully delivered roughly 830 and 825 packets, corresponding to PLRs of approximately 7.8\% and 8.3\% (average $\approx$8\%). Given the 30-min sampling interval, this loss rate is operationally acceptable and leaves negligible gaps in canopy monitoring over the growth stage.

Together, the two-unit deployment supports the operational feasibility of integrating the LCA-enhanced four-class model into a solar-powered edge system for repeated canopy defoliation assessment and remote monitoring in grapevines. The camera system also demonstrated robust thermal tolerance and stability, maintaining reliable operation on solar power even when internal enclosure temperatures reached approximately 35\,$^{\circ}$C for 6--7\,h per day. Furthermore, the battery level remained above 75\% on average each day, confirming its practicality for continuous, unattended field deployment.

\begin{table*}[!t]
\centering
\caption{Comparison of DefoliationCAM with existing deep-learning approaches
for plant defoliation, disease, and pest severity assessment in terms of
model, dataset, task, platform, camera vision, and performance.}
\label{tab:comparison}
\resizebox{\textwidth}{!}{%
\begin{tabular}{@{} >{\raggedright\arraybackslash}p{2.7cm} >{\raggedright\arraybackslash}p{3.0cm} >{\raggedright\arraybackslash}p{4.2cm} >{\raggedright\arraybackslash}p{3.2cm} >{\raggedright\arraybackslash}p{2.4cm} >{\raggedright\arraybackslash}p{1.9cm} >{\raggedright\arraybackslash}p{3.3cm} @{}}
\toprule
\textbf{Study} & \textbf{Model} & \textbf{Dataset} & \textbf{Assessment Task} & \textbf{Platform} & \textbf{Camera Vision} & \textbf{Performance} \\
\midrule
\textbf{DefoliationCAM} & YOLOX-Nano+LCA-enhanced MobileNet classifier (1.08\,M) &
Field-collected dataset of 29{,}674 field-collected grape canopy patches with AI-generated damaged-class augmentation &
4-class defoliation severity (background, no to negligible damage, mild, moderate-to-severe) aggregated into NGDS score &
NVIDIA Jetson Orin Nano (TensorRT FP16) &
Field-scale (4K canopy) &
83.6\% Acc, 0.838 macro-F1;
NGDS mapped to three QDA-derived management categories (boundaries at 9\% and 29\%);
$\sim$1.5\,s per 4K image \\
\midrule
Lazcano-Garc\'ia et al.~(2025)~\cite{lazcano2025} & YOLOv5 &
3{,}198 grape leaf images (800 molecularly diagnosed) &
2-class virus symptoms (Red Blotch, Leafroll) &
NVIDIA Jetson Nano &
Near-view (leaf) &
95.36\% Precision, 95.56\% F1-score \\
\midrule
Karim et al.~(2024)~\cite{karim2024} & Modified MobileNetV3-Large + Grad-CAM &
Study-specific grape leaf disease image set &
4-class grape leaf disease &
NVIDIA Jetson Nano &
Near-view (leaf) &
99.42\% test accuracy; 3.4\,W avg.\ power \\
\midrule
Xie et al.~(2020)~\cite{xie2020} & Faster DR-IACNN (Faster R-CNN + Inception + SE) &
Study-specific GLDD grape leaf disease dataset &
4-class disease detection (Black rot, Black measles, Leaf blight, Mites) &
Server &
Near-view (leaf) &
81.1\% mAP, 15.01\,FPS \\
\midrule
Ilyas et al.~(2022)~\cite{ilyas2022} & DIANA (region-based hybrid CNN framework) &
Study-specific paprika disease dataset (6 diseases $\times$ 11 severity stages) &
Multi-disease severity phenotyping with descriptive phrases &
Server &
Near-view (single plant) &
End-to-end severity localization and description \\
\midrule
Sun et al.~(2025)~\cite{sun2025} & YOLO-PLNet (lightweight YOLO11n, 2.13\,M) &
Study-specific peanut leaf disease image set &
Multi-class peanut leaf disease detection &
NVIDIA Jetson Orin NX (TensorRT FP16/INT8) &
Near-view (leaf) &
98.1\% mAP@0.5; 28.2\,FPS (FP16), 41.3\,FPS (INT8) \\
\bottomrule
\end{tabular}%
}
\end{table*}

\subsection{Comparison with Previous Studies}

Few studies have directly addressed the automated assessment of pest-induced defoliation. The comparison was therefore broadened to include plant disease and pest studies addressing related image-classification, object-detection, and severity-assessment tasks (Table~\ref{tab:comparison}). Although the targets are different, these studies provide relevant technical context for comparing model architecture, imaging scale, computing platform, and deployment capability. 

Xie et~al.~\cite{xie2020} applied a Faster R-CNN-based model to disease detection in close-up grape leaf images, whereas DIANA~\cite{ilyas2022} performed disease and pest severity phenotyping on individual paprika plants. Both systems were evaluated using server-based processing and did not address autonomous, canopy-scale field monitoring.

Edge deployment has also been investigated for grapevines. Karim et~al.~\cite{karim2024} implemented a modified MobileNetV3-Large model on a Jetson Nano for four-class grape leaf disease classification through a graphical user interface. Lazcano-Garc'ia et~al.~\cite{lazcano2025} similarly deployed YOLOv5 on a Jetson Nano to identify grapevine red blotch and leafroll virus symptoms. For another crop, Sun et~al.~\cite{sun2025} deployed YOLO-PLNet on a Jetson Orin~NX for real-time peanut leaf disease detection using an industrial CSI camera. The system achieved high processing rates under TensorRT FP16 and INT8 inference, demonstrating the computational efficiency of lightweight models on edge hardware.  These studies demonstrated grape-specific inference on embedded hardware but relied on near-view leaf images and operator-initiated image acquisition rather than scheduled canopy monitoring. Solar-powered, unattended field operation was not evaluated.

D-CAM addresses a different operational objective by combining field-scale 4K grape-canopy imaging, on-device cordon detection and four-class patch classification, scheduled solar-powered operation, and LoRa-based transmission of summary results. Patch predictions are aggregated into a canopy-level NGDS, which is related to three management categories through QDA-derived boundaries based on entomologist assignments. Among the systems summarized in Table~\ref{tab:comparison}, D-CAM is distinguished by its integration of canopy-scale assessment, autonomous edge inference, battery monitoring, and multi-week vineyard deployment.

\section{Conclusion}
This study developed DefoliationCAM (D-CAM), a solar-powered edge AI system for autonomous assessment of grape canopy defoliation. The two-stage pipeline combined YOLOX-Nano canopy detection with an LCA-enhanced patch classifier and converted the patch predictions into a canopy-level Non-linear Global Defoliation Severity (NGDS) score. Results were transmitted through LoRa without requiring direct Internet access at each field unit. Consolidating the moderate and severe classes into a single moderate-to-severe damage class, followed by synthetic augmentation, increased macro-F1 from 0.685 to 0.838. The four-class model with nonoverlapping tiling also achieved closer agreement with the entomologist-derived ground-truth NGDS than the five-class and overlapping-tiling configurations. The LCA-enhanced classifier achieved higher classification performance than ResNet-50 despite having substantially fewer parameters (1.08 M versus 23.52 M).

Image-level evaluation was conducted using 86 held-out canopy images. Patch labels assigned by an entomologist were used to calculate the ground-truth NGDS for each image, and the entomologist independently assigned each complete canopy image to a management category. Quadratic discriminant analysis identified NGDS boundaries of 9\% and 29\%, defining No to negligible damage, Mild-damage, and Action-level-damage categories. A multi-week deployment of two field units further demonstrated autonomous image acquisition, edge inference, LoRa communication, battery monitoring, and remote visualization under vineyard conditions. The recorded NGDS trends were consistent with the field-observed accumulation of Japanese beetle feeding damage.

Future work will focus on improving image quality control and extending the system to mobile robotic platforms. First, a lightweight post-capture screening module can be developed to identify images affected by poor focus, unsuitable exposure, or excessive canopy occlusion. Images that do not meet the quality criteria could be recaptured or excluded before inference. Second, transfer learning will be used to adapt the current detector and classifier to side-view canopy images collected by a ground robot or unmanned ground vehicle. The existing models will be fine-tuned using a limited labeled set of side-view images, which may reveal defoliation that is less visible from the current fixed-camera perspective while reducing additional annotation requirements. View-specific calibration will also be evaluated before NGDS estimates from different imaging perspectives are compared or combined.

\section*{Acknowledgments}
The authors gratefully acknowledge the support of the Richard P. Kimmel and Laurine Kimmel Charitable Foundation, U.S. National Science Foundation Experiential Learning for Emerging and Novel Technologies (ExLENT) Program (Award No. 2322535), USDA-CBG (Award No. 2024-38821-42099), USDA-NIFA (Award No. 2023-69017-40560), and NSF-EPSCoR (Award No. 2420405).

\end{document}